\documentclass{article}

\usepackage[final]{nips_2017}

\usepackage[utf8]{inputenc} % allow utf-8 input
\usepackage[T1]{fontenc}    % use 8-bit T1 fonts
\usepackage{hyperref}       % hyperlinks
\usepackage{url}            % simple URL typesetting
\usepackage{booktabs}       % professional-quality tables
\usepackage{amsfonts}       % blackboard math symbols
\usepackage{nicefrac}       % compact symbols for 1/2, etc.
\usepackage{microtype}      % microtypography
\usepackage{amsmath,amssymb,amsthm}
\usepackage{mathtools}
\usepackage{color}
\usepackage{algorithm}
\usepackage{algorithmic}
\usepackage{graphicx}
\usepackage{slashbox}
\usepackage[table]{xcolor}
\usepackage{multirow}
\usepackage{rotating}
\usepackage{caption}
\usepackage{enumitem}

\setlist[itemize]{leftmargin=*}

\DeclareMathOperator*{\argmax}{arg\!\max}
\DeclareMathOperator*{\argmin}{arg\!\min}

\DeclarePairedDelimiter\ceil{\lceil}{\rceil}
\DeclarePairedDelimiter\floor{\lfloor}{\rfloor}

\newcommand{\prob}[1]{{\mathbb{P}\left(#1\right)}}

\newcommand{\RB}{RB}
\newcommand{\MAB}{MAB}
\newcommand{\CTO}{CTO}
\newcommand{\SWA}{SWA}
\newcommand{\wSWA}{wSWA}
\newcommand{\NPC}{NPC}
\newcommand{\PC}{PC}
\newcommand{\AV}{AV}
\newcommand{\ANV}{ANV}

\newtheorem{theorem}{Theorem}[section]
\newtheorem{corollary}{Corollary}[theorem]
\newtheorem{definition}{Definition}[section]
\newtheorem{assumption}{Assumption}[section]

\newtheorem{lemma}{Lemma}[section]

\newtheorem{example}{Example}[section]

\title{Rotting Bandits}

\author{
  Nir Levine \\
  Electrical Engineering Department\\
  The Technion\\
  Haifa 32000, Israel \\
  \texttt{levin.nir1@gmail.com} \\
  \And
  Koby Crammer \\
  Electrical Engineering Department\\
  The Technion \\
  Haifa 32000, Israel \\
  \texttt{koby@ee.technion.ac.il} \\
  \AND
  Shie Mannor \\
  Electrical Engineering Department\\
  The Technion \\
  Haifa 32000, Israel \\
  \texttt{shie@ee.technion.ac.il} \\
}

\begin{document}

\maketitle

\begin{abstract} 
The Multi-Armed Bandits (MAB) framework highlights the trade-off between acquiring new knowledge (Exploration) and leveraging available knowledge (Exploitation). In the classical MAB problem, a decision maker must choose an arm at each time step, upon which she receives a reward. The decision maker's objective is to maximize her cumulative expected reward over the time horizon. The MAB problem has been studied extensively, specifically under the assumption of the arms' rewards distributions being stationary, or quasi-stationary, over time. We consider a variant of the MAB framework, which we termed \textit{Rotting Bandits}, where each arm's expected reward decays as a function of the number of times it has been pulled. We are motivated by many real-world scenarios such as online advertising, content recommendation, crowdsourcing, and more. We present algorithms, accompanied by simulations, and derive theoretical guarantees.
\end{abstract} 

\section{Introduction}
%% MAB in general
% explain in general about MAB & regret(e.g, originally from Thompson then Robbins 1952)
One of the most fundamental trade-offs in stochastic decision theory is the well celebrated Exploration vs. Exploitation dilemma. Should one acquire new knowledge on the expense of possible sacrifice in the immediate reward (Exploration), or leverage past knowledge in order to maximize instantaneous reward (Exploitation)? Solutions that have been demonstrated to perform well are those which succeed in balancing the two. First proposed by~\citet{ThompsonLikelihood} in the context of drug trials, and later formulated in a more general setting by~\citet{SeqDesignExp}, MAB problems serve as a distilled framework for this dilemma. In the classical setting of the MAB, at each time step, the decision maker must choose (pull) between a fixed number of arms. After pulling an arm, she receives a reward which is a realization drawn from the arm's underlying reward distribution. The decision maker's objective is to maximize her cumulative expected reward over the time horizon. An equivalent, more typically studied, is the \textit{regret}, which is defined as the difference between the optimal cumulative expected reward (under full information) and that of the policy deployed by the decision maker.

% examples for uses of MAB; ..but not limited to.. (e.g., online advertising (taxonomies), routing (addpative routing), online auctions (the value of knowing the demand curve))
MAB formulation has been studied extensively, and was leveraged to formulate many real-world problems. Some examples for such modeling are online advertising \citep{BanditsTaxonomies}, routing of packets \citep{AdaptiveRouting}, and online auctions \citep{ValueKnowingDemand}.

% our motivation & our problem's model & policy regert (opt is mixed actions.. e.g., from regret to policy regret)
% (e.g., saptio-temporal, efficient crowdsourcing)
Most past work (Section~\ref{sec_Rel_work}) on the MAB framework has been performed under the assumption that the underlying distributions are stationary, or possibly quasi-stationary. In many real-world scenarios, this assumption may seem simplistic. Specifically, we are motivated by real-world scenarios where the expected reward of an arm decreases over time instances that it has been pulled. We term this variant \textit{Rotting Bandits}. For motivational purposes, we present the following two examples.
\begin{itemize}
	\item Consider an online advertising problem where an agent must choose which ad (arm) to present (pull) to a user. It seems reasonable that the effectiveness (reward) of a specific ad on a user would deteriorate over exposures. Similarly, in the content recommendation context, \citet{SpatioTemporal} showed that articles' CTR decay over amount of exposures.
	\item Consider the problem of assigning projects through crowdsourcing systems \citep{EfficientCrowd}. Given that the assignments primarily require human perception, subjects may fall into boredom and their performance would decay (e.g., license plate transcriptions \citep{AutoLicPlate}).
\end{itemize}

As opposed to the stationary case, where the optimal policy is to always choose some specific arm, in the case of Rotting Bandits the optimal policy consists of choosing different arms. This results in the notion of \textit{adversarial regret} vs. \textit{policy regret} \citep{RegretToPolicyRegret} (see Section~\ref{sec_Rel_work}). In this work we tackle the harder problem of minimizing the policy regret.

The main contributions of this paper are the following:
\begin{itemize}
	\item Introducing a novel, real-world oriented \MAB\ formulation, termed \textit{Rotting Bandits}.
	\item Present an easy-to-follow algorithm for the general case, accompanied with theoretical guarantees.
	\item Refine the theoretical guarantees for the case of existing prior knowledge on the rotting models, accompanied with suitable algorithms.
\end{itemize}

% paper construction
The rest of the paper is organized as follows: in Section~\ref{sec_model} we present the model and relevant preliminaries. In Section~\ref{sec_non_param} we present our algorithm along with theoretical guarantees for the general case. In Section~\ref{sec_param} we do the same for the parameterized case, followed by simulations in Section~\ref{sec_sim}. In Section~\ref{sec_Rel_work} we review related work, and conclude with a discussion in Section~\ref{Sec_dis}.

\section{Model and Preliminaries}
\label{sec_model}
We consider the problem of Rotting Bandits (\RB); an agent is given $K$ arms and at each time step $t=1,2,..$ one of the arms must be pulled. We denote the arm that is pulled at time step $t$ as ${i\left(t\right) \in \left[K\right]=\{1,..,K\}}$. When arm $i$ is pulled for the $n^{\textrm{th}}$ time, the agent receives a time independent, $\sigma^{2}$ sub-Gaussian random reward, $r_{t}$, with mean ${\mu_{i}\left(n\right)}$.\footnote{Our results hold for pulls-number dependent variances $\sigma^{2}\left(n\right)$, by upper bound them ${\sigma^{2} \geq \sigma^{2}\left(n\right), \forall n}$. It is fairly straightforward to adapt the results to pulls-number dependent variances, but we believe that the way presented conveys the setting in the clearest way.}

In this work we consider two cases: (1) There is no prior knowledge on the expected rewards, except for the `rotting' assumption to be presented shortly, i.e., a non-parametric case (\NPC). (2) There is prior knowledge that the expected rewards comprised of an unknown constant part and a \textit{rotting} part which is known to belong to a set of rotting models, i.e., a parametric case (\PC).

Let $N_{i}\left(t\right)$ be the number of pulls of arm $i$ at time $t$ not including this round's choice (${N_{i}\left(1\right)=0}$), and $\Pi$ the set of all sequences ${i\left(1\right), i\left(2\right),..}$, where ${i\left(t\right) \in \left[K\right], \forall t \in \mathbb{N}}$. i.e., ${\pi \in \Pi}$ is an infinite sequence of actions (arms), also referred to as a policy. We denote the arm that is chosen by policy $\pi$ at time $t$ as $\pi\left(t\right)$. The objective of an agent is to maximize the expected total reward in time $T$, defined for policy $\pi \in \Pi$ by,
\begin{equation}
J\left(T ; \pi\right) = \mathbb{E}\left[\sum_{t=1}^{T} \mu_{\pi\left(t\right)}\left(N_{\pi\left(t\right)}\left(t\right)+1\right) \right]
\label{total_reward}
\end{equation}
We consider the equivalent objective of minimizing the regret in time $T$ defined by,
\begin{equation}
\mathcal{R}\left(T ; \pi\right) = \max_{\tilde{\pi} \in \Pi} \{ J\left(T ; \tilde{\pi}\right) \} - J\left(T ; \pi\right).
\label{regret}
\end{equation}

\begin{assumption}
\textbf{(Rotting)} $\forall i \in \left[K\right]$, $\mu_{i}\left(n\right)$ is positive, and non-increasing in $n$.
\label{assm_rotting}
\end{assumption}

\subsection{Optimal Policy}
Let $\pi^{\textrm{max}}$ be a policy defined by,
\begin{equation}
\pi^{\textrm{max}}\left(t\right) \in \argmax_{i\in\left[K\right]} \{\mu_{i}\left(N_{i}\left(t\right)+1\right)\}
\label{opt_policy}
\end{equation}
where, in a case of tie, break it randomly.

\begin{lemma}
\label{lemma_opt_policy}
$\pi^{\normalfont{\textrm{max}}}$ is an optimal policy for the \RB\ problem. \\
\textbf{Proof:} See Appendix B of the supplementary material.
\end{lemma}

\section{Non-Parametric Case}
\label{sec_non_param}
In the \NPC\ setting for the \RB\ problem, the only information we have is that the expected rewards sequences are positive and non-increasing in the number of pulls. The Sliding-Window Average (\textbf{\SWA}) approach is a heuristic for ensuring with high probability that, at each time step, the agent did not sample significantly sub-optimal arms too many times. We note that, potentially, the optimal arm changes throughout the trajectory, as Lemma~\ref{lemma_opt_policy} suggests. We start by assuming that we know the time horizon, and later account for the case we do not.

\underline{\textbf{Known Horizon}}\\
The idea behind the \SWA\ approach is that after we pulled a significantly sub-optimal arm ``enough" times, the empirical average of these ``enough" pulls would be distinguishable from the optimal arm for that time step and, as such, given any time step there is a bounded number of significantly sub-optimal pulls compared to the optimal policy. Pseudo algorithm for \SWA\ is given by Algorithm~\ref{alg_swa}.

\begin{algorithm}
\caption{\SWA} \label{alg_swa}
\begin{algorithmic}
\STATE $\textbf{Input}: K, T, \alpha > 0$
\STATE $\textbf{Initialize}: M \leftarrow \ceil{\alpha 4^{2/3} \sigma^{2/3} K^{-2/3} T^{2/3} \ln ^{1/3} \left(\sqrt{2} T \right)} \text{, and } N_{i} \leftarrow 0 \text{ for all } i \in \left[K\right]$
\FOR {$t = 1,2,..,KM$}
\STATE $\textbf{Ramp up}: i\left(t\right) \text{by Round-Robin, receive } r_{t} \text{, and set } N_{i\left(t\right)} \leftarrow N_{i\left(t\right)} + 1 \text{ ; } r_{i\left(t\right)}^{N_{i\left(t\right)}} \leftarrow r_{t}$
\ENDFOR
\FOR {$t = KM+1,...,T$}
\STATE $\textbf{Balance}: i\left(t\right) \in \argmax_{i\in\left[K\right]} \bigg\{ \frac{1}{M} \sum_{n=N_{i}-M+1}^{N_{i}} r_{i}^{n} \bigg\}$
\STATE $\textbf{Update}: \text{receive } r_{t} \text{, and set } N_{i\left(t\right)} \leftarrow N_{i\left(t\right)} + 1 \text{ ; } r_{i\left(t\right)}^{N_{i\left(t\right)}} \leftarrow r_{t}$
\ENDFOR
\end{algorithmic}
\end{algorithm}

\begin{theorem}
Suppose Assumption~\ref{assm_rotting} holds. \SWA\ algorithm achieves regret bounded by,
\begin{equation}
\mathcal{R}\left(T ; \pi^{\normalfont{\textrm{\SWA}}}\right) \leq \left(\alpha \max_{i \in \left[K\right]} \mu_{i}\left(1\right) + \alpha^{-1/2}\right) 4^{2/3} \sigma^{2/3} K^{1/3} T^{2/3} \ln ^{1/3} \left(\sqrt{2} T \right) + 3K \max_{i \in \left[K\right]} \mu_{i}\left(1\right)
\end{equation}
\label{thm_swa_known_horizon}
\textbf{Proof:} See Appendix C.1 of the supplementary material.
\end{theorem}
We note that the upper bound obtains its minimum for $\alpha = \left(2 \max_{i \in \left[K\right]} \mu_{i}\left(1\right)\right)^{-2/3}$, which can serve as a way to choose $\alpha$ if $\max_{i \in \left[K\right]} \mu_{i}\left(1\right)$ is known, but $\alpha$ can also be given as an input to \SWA\ to allow control on the averaging window size.

\underline{\textbf{Unknown Horizon}}\\
In this case we use \textit{doubling trick} in order to achieve the same horizon-dependent rate for the regret. We apply the \SWA\ algorithm with a series of increasing horizons (powers of two, i.e., $1,2,4,..$) until reaching the (unknown) horizon. We term this Algorithm \wSWA\ (wrapper \SWA).

\begin{corollary}
Suppose Assumption~\ref{assm_rotting} holds. \wSWA\ algorithm achieves regret bounded by,
\begin{multline}
\mathcal{R}\left(T ; \pi^{\normalfont{\textrm{\wSWA}}}\right) \leq \left(\alpha \max_{i \in \left[K\right]} \mu_{i}\left(1\right) + \alpha^{-1/2}\right) 8 \sigma^{2/3} K^{1/3} T^{2/3} \ln^{1/3} \left(\sqrt{2}T \right) \\ + 3K \max_{i \in \left[K\right]} \mu_{i}\left(1\right) \left( \log_{2}T+1 \right)
\end{multline}
\label{crl_swa_unknown_horizon}
\textbf{Proof:} See Appendix C.2 of the supplementary material.
\end{corollary}

\section{Parametric Case}
\label{sec_param}
In the \PC\ setting for the \RB\ problem, there is prior knowledge that the expected rewards comprised of a sum of an unknown constant part and a rotting part known to belong to a set of models, $\Theta$. i.e., the expected reward of arm $i$ at its $n^{\textrm{th}}$ pull is given by, ${\mu_{i}\left(n\right) = \mu_{i}^{c} + \mu\left(n ; \theta_{i}^{*} \right)}$, where ${\theta_{i}^{*} \in \Theta}$. We denote ${\{\theta_{i}^{*}\}_{i=1}^{\left[K\right]}}$ by $\Theta^{*}$. We consider two cases: The first is the asymptotically vanishing case (\AV), i.e., ${\forall i : \mu_{i}^{c}=0}$. The second is the asymptotically non-vanishing case (\ANV), i.e., ${\forall i : \mu_{i}^{c} \in \mathbb{R}}$.

We present a few definitions that will serve us in the following section.

\begin{definition}
For a function ${f : \mathbb{N} \rightarrow \mathbb{R}}$, we define the function ${f^{\star \downarrow} : \mathbb{R} \rightarrow \mathbb{N} \cup \{\infty\}}$ by the following rule: given ${\zeta \in \mathbb{R}}$, ${f^{\star \downarrow}\left(\zeta\right)}$ returns the smallest ${N \in \mathbb{N}}$ such that ${\forall n \geq N : f\left(n\right) \leq \zeta}$, or $\infty$ if such $N$ does not exist.
\end{definition}

\begin{definition} For any $\theta_{1} \neq \theta_{2} \in \Theta^{2}$, define $det_{\theta_{1}, \theta_{2}}, Ddet_{\theta_{1}, \theta_{2}} : \mathbb{N} \rightarrow \mathbb{R}$ as,
\begin{equation} \nonumber
\begin{split}
det_{\theta_{1}, \theta_{2}}\left(n\right) & = \frac{n \sigma^{2}}{\left( \sum_{j=1}^{n} \mu\left(j ; \theta_{1}\right) - \sum_{j=1}^{n} \mu\left(j ; \theta_{2}\right) \right)^{2}} \\
Ddet_{\theta_{1}, \theta_{2}}\left(n\right) & = \frac{n \sigma^{2}}{\left( \sum_{j=1}^{\floor{n/2}} \left[ \mu\left(j ; \theta_{1}\right) - \mu\left(j ; \theta_{2}\right) \right] - \sum_{j=\floor{n/2}+1}^{n} \left[ \mu\left(j ; \theta_{1}\right) - \mu\left(j ; \theta_{2}\right) \right] \right)^{2}}
\end{split}
\end{equation}
\end{definition}

\begin{definition}
Let ${bal : \mathbb{N} \cup \infty \rightarrow \mathbb{N} \cup \infty}$ be defined at each point $n \in \mathbb{N}$ as the solution for,
\begin{equation} \nonumber
\begin{split}
\min \text{ } \alpha \qquad
\normalfont \text{s.t, } 
\max_{\theta \in \Theta} \mu\left(\alpha ; \theta\right) \leq \min_{\theta \in \Theta} \mu\left(n ; \theta\right)
\end{split}
\end{equation}
We define $bal\left(\infty\right) = \infty$.
\end{definition}

\begin{assumption}
\textbf{(Rotting Models)} $\mu\left(n ; \theta\right)$ is positive, non-increasing in $n$, and ${\mu\left(n ; \theta\right) \in o\left(1\right)}$, ${\forall \theta \in \Theta}$, where $\Theta$ is a discrete known set.
\label{assm_param_rotting}
\end{assumption}

We present an example for which, in Appendix~E, we demonstrate how the different following assumptions hold. By this we intend 
to achieve two things: (i) show that the assumptions are not too harsh, keeping the problem relevant and non-trivial, and (ii) present a simple example on how to verify the assumptions.

\begin{example}
The reward of arm $i$ for its $n^{\textrm{th}}$ pull is distributed as ${\mathcal{N}\left(\mu_{i}^{c}+n^{-\theta_{i}^{*}}, \sigma^{2}\right)}$. Where ${\theta_{i}^{*} \in \Theta = \{\theta_{1},\theta_{2},...,\theta_{M}\}}$, and ${\forall \theta \in \Theta : 0.01 \leq\theta \leq 0.49}$.
\label{expl}
\end{example}

\subsection{Closest To Origin (\AV)}
The Closest To Origin (\CTO) approach for \RB\ is a heuristic that simply states that we hypothesize that the true underlying model for an arm is the one that best fits the past rewards. The fitting criterion is proximity to the origin of the sum of expected rewards shifted by the observed rewards. Let ${r^{i}_{1}, r^{i}_{2},.., r^{i}_{N_{i}\left(t\right)}}$ be the sequence of rewards observed from arm $i$ up until time $t$. Define,
\begin{equation}
Y\left(i, t ; \Theta\right) = \bigg\{ \sum_{j=1}^{N_{i}\left(t\right)} r^{i}_{j} - \sum_{j=1}^{N_{i}\left(t\right)} \mu\left(j ; \theta\right) \bigg\}_{\theta \in \Theta}.
\label{wa_set}
\end{equation}
The \CTO\ approach dictates that at each decision point, we assume that the true underlying rotting model corresponds to the following proximity to origin rule (hence the name),
\begin{equation}
\hat{\theta}_{i}\left(t\right) = \argmin_{\theta \in \Theta} \{| Y\left(i,t ; \theta\right)| \}.
\label{theta_choose}
\end{equation}
The \CTO$_{\textrm{SIM}}$ version tackles the \RB\ problem by simultaneously detecting the true rotting models and balancing between the expected rewards (following Lemma~\ref{lemma_opt_policy}). In this approach, every time step, each arm's rotting model is hypothesized according to the proximity rule~(\ref{theta_choose}). Then the algorithm simply follows an $\argmax$ rule, where least number of pulls is used for tie breaking (randomly between an equal number of pulls). Pseudo algorithm for \CTO$_{\textrm{SIM}}$ is given by Algorithm~\ref{cto_sim_pure_alg}.

\begin{assumption}
\textbf{(Simultaneous Balance and Detection ability)} 
\begin{equation} \nonumber
bal\left(\max_{\theta_{1} \neq \theta_{2} \in \Theta^{2}} \bigg\{det_{\theta_{1}, \theta_{2}}^{\star \downarrow}\left(\frac{1}{16}  \ln^{-1}\left(\zeta\right)\right)\bigg\}\right) \in o \left(\zeta\right)
\end{equation}
\label{assm_sim}
\end{assumption}

The above assumption ensures that, starting from some horizon $T$, the underlying models could be distinguished from the others, w.p $1-1/T^{2}$, by their sums of expected rewards, and the arms could then be balanced, all within the horizon.

\begin{theorem}
Suppose Assumptions~\ref{assm_param_rotting} and \ref{assm_sim} hold. There exists a finite step ${T^{*}_{\normalfont{\textrm{SIM}}}}$, such that for all ${T \geq T^{*}_{\normalfont{\textrm{SIM}}}}$, \textnormal{\CTO}$_{\normalfont{\textrm{SIM}}}$ achieves regret upper bounded by $\mathbf{o\left(1\right)}$ (which is upper bounded by $\max_{\theta \in \Theta^{*}} \mu\left(1;\theta\right)$). Furthermore, ${T^{*}_{\normalfont{\textrm{SIM}}}}$ is upper bounded by the solution for the following,
\begin{equation}
\begin{split}
\min & \text{ } T \\
\text{s.t } &  \begin{cases} 
T,b \in \mathbb{N}\cup{\{0\}}, t \in \mathbb{N}^{K} \\
\forall b, \exists t : \begin{cases}
\|t\|_{1} \leq T+b \\
t_{i} \geq \max_{\theta \in \Theta^{*}}{\bigg\{m^{*}\left(\frac{1}{K\left(T+b\right)^{2}} ; \theta\right)\bigg\}} \\
\mu\left(t_{i}+1 ; \theta^{*}_{i}\right) \leq  \min_{\tilde{\theta} \in \Theta} \left[\mu\left(\max_{\theta \in \Theta^{*}}{\bigg\{m^{*}\left(\frac{1}{K\left(T+b\right)^{2}} ; \theta\right)\bigg\}} ; \tilde{\theta}\right) \right]
\end{cases}
\end{cases}
\end{split}
\label{T_SIM_upper_bound}
\end{equation}
\label{thm_cto_sim_pure}
\textbf{Proof}: See Appendix D.1 of the supplementary material.
\end{theorem}

Regret upper bounded by $o\left(1\right)$ is achieved by proving that w.p of $1-1/T$ the regret vanishes, and in any case it is still bounded by a decaying term. The shown optimization bound stems from ensuring that the arms would be pulled enough times to be correctly detected, and then balanced (following the optimal policy, Lemma~\ref{lemma_opt_policy}). Another upper bound for ${T^{*}_{\normalfont{\textrm{SIM}}}}$ can be found in Appendix~D.1.

\subsection{Differences Closest To Origin (\ANV)}
We tackle this problem by estimating both the rotting models and the constant terms of the arms. The Differences Closest To Origin (\textbf{D-\CTO}) approach is composed of two stages: first, detecting the underlying rotting models, then estimating and controlling the pulls due to the constant terms. We denote ${a^{*} = \argmax_{i \in \left[K\right]} \{\mu_{i}^{c}\}}$, and $\Delta_{i} = \mu_{a^{*}}^{c} - \mu_{i}^{c}$.

\begin{assumption}
\textbf{(D-Detection ability)} 
\begin{equation} \nonumber
\max_{\theta_{1} \neq \theta_{2} \in \Theta^{2}} \bigg\{Ddet^{\star \downarrow}_{\theta_{1}, \theta_{2}}\left(\epsilon\right)\bigg\} \leq D\left(\epsilon\right) < \infty, \quad \forall \epsilon > 0
\end{equation}
\label{assm_Bdistinction}
\vspace{-0.01in}
\end{assumption}
This assumption ensures that for any given probability, the models could be distinguished, by the differences (in pulls) between the first and second halves of the models' sums of expected rewards.

\underline{\textbf{Models Detection}}\\
In order to detect the underlying rotting models, we cancel the influence of the constant terms. Once we do this, we can detect the underlying models. Specifically, we define a criterion of proximity to the origin based on differences between the halves of the rewards sequences, as follows: define,
\begin{equation}
Z\left(i, t ; \Theta\right) = \left(\sum_{j=1}^{\floor*{N_{i}\left(t\right)/2}}r_{j}^{i} - \sum_{j=\floor*{N_{i}\left(t\right)/2}+1}^{N_{i}\left(t\right)}r_{j}^{i}\right) - \left(\sum_{j=1}^{\floor*{N_{i}\left(t\right)/2}}\mu\left(j ; \theta\right) - \sum_{j=\floor*{N_{i}\left(t\right)/2}+1}^{N_{i}\left(t\right)}\mu\left(j ; \theta\right)\right).
\label{wa_bias_set}
\end{equation}
The D-\CTO\ approach is that in each decision point, we assume that the true underlying model corresponds to the following rule,
\begin{equation}
\hat{\theta}_{i}\left(t\right) = \argmin_{\theta \in \Theta} \{ | Z\left(i, t ; \theta\right) | \}
\label{theta_bias_choose}
\end{equation}
We define the following optimization problem, indicating the number of samples required for ensuring correct detection of the rotting models w.h.p. For some arm $i$ with (unknown) rotting model $\theta_{i}^{*}$,
\begin{equation}
\begin{split}
\min \text{ } m  \qquad
\text{s.t } \begin{cases} 
P\left(\hat{\theta}_{i}\left(l\right) \neq \theta_{i}^{*} \right) \leq p, \quad \forall l \geq m \\
\text{while pulling only arm } i.
\end{cases}
\end{split}
\label{opt_prob_dcto}
\end{equation}
We denote the solution to the above problem, when we use proximity rule~(\ref{theta_bias_choose}), by $m^{*}_{\textrm{diff}}\left(p ; \theta_{i}^{*}\right)$, and define ${m^{*}_{\textrm{diff}}\left(p\right) = \max_{\theta \in \Theta}{\{m^{*}_{\textrm{diff}}\left(p ; \theta\right)\}}}$. 

\pagebreak

\underline{\textbf{D-\CTO$_{\textrm{UCB}}$}}\\
We next describe an approach with one decision point, and later on remark on the possibility of having a decision point at each time step.
As explained above, after detecting the rotting models, we move to tackle the constant terms aspect of the expected rewards. This is done in a UCB1-like approach \citep{FiniteTimeAnalysisMAB}. Given a sequence of rewards from arm $i$, ${\{r_{k}^{i}\}_{k=1}^{N_{i}\left(t\right)}}$, we modify them using the estimated rotting model $\hat{\theta}_{i}$, then estimate the arm's constant term, and finally choose the arm with the highest estimated expected reward, plus an upper confident term. i.e., at time $t$, we pull arm $i\left(t\right)$, according to the rule,
\begin{equation}
i\left(t\right) \in \argmax_{i \in \left[K\right]} \left[\hat{\mu}_{i}^{c}\left(t\right) + \mu\left(N_{i}\left(t\right)+1 ; \hat{\theta}_{i}\left(t\right)\right) + c_{t,N_{i}\left(t\right)} \right]
\label{ucb_arm}
\end{equation}
where $\hat{\theta}_{i}\left(t\right)$ is the estimated rotting model (obtained in the first stage), and,
\begin{equation} \nonumber
\hat{\mu}_{i}^{c}\left(t\right) = \frac{\sum_{j=1}^{N_{i}\left(t\right)}\left(r_{j}^{i} - \mu\left(j ; \hat{\theta}_{i}\left(t\right)\right)\right)}{N_{i}\left(t\right)} , \qquad
c_{t,s} = \sqrt{\frac{8 \ln \left(t\right) \sigma^{2}}{s}}
\end{equation}
In a case of a tie in the UCB step, it may be arbitrarily broken. Pseudo algorithm for D-\CTO$_{\textrm{UCB}}$ is given by Algorithm~\ref{bcto_alg}, accompanied with the following theorem.

\begin{theorem}
Suppose Assumptions~\ref{assm_param_rotting}, and \ref{assm_Bdistinction} hold. For $\delta \in\left(0,1\right)$, with probability of at least $1-\delta$, \textnormal{D-\CTO}$_{\normalfont{\textrm{UCB}}}$ algorithm  achieves regret bounded at time $T$ by,
\begin{equation}
\sum_{\substack{i \in \left[K\right]\\ i \neq a^{*}}} \bigg[\max \bigg\{ m^{*}_{\normalfont{\textrm{diff}}}\left(\delta/K\right), \mu^{\star \downarrow}\left(\epsilon_{i} ; \theta_{i}^{*}\right), \frac{32 \sigma^{2} \ln T}{\left(\Delta_{i} - \epsilon_{i}\right)^{2}} \bigg\} 
\times \left(\Delta_{i} + \mu\left(1 ; \theta_{a^{*}}^{*}\right)\right)\bigg] + C\left(\Theta^{*}, \{\mu_{i}^{c}\}\right)
\label{opt_prob_bcto_regret}
\end{equation}
for any sequence $\epsilon_{i} \in \left(0, \Delta_{i}\right), \forall i \neq a^{*}$. Where ${\frac{32 \sigma^{2} \ln T}{\left(\Delta_{i} - \epsilon_{i}\right)^{2}}}$ is the only time-dependent factor.
\label{thm_Bcto}
\\
\textbf{Proof}: See Appendix D.2 of the supplementary material.
\end{theorem}

\begin{table}[t]
\begin{tabular}{c|c}
\begin{minipage}{.54\textwidth}
\centering
\begin{algorithm}[H]
\caption{\CTO$_{\textnormal{SIM}}$}
\label{cto_sim_pure_alg}
\begin{algorithmic}
\STATE $\textbf{Input}: K, \Theta$
\STATE $\textbf{Initialization}: N_{i}=0, \text{ } \forall i \in \left[K\right]$
\FOR {$t = 1,2,..,K$}
\STATE $\textbf{Ramp up}: i\left(t\right)=t \text{ ,and update } N_{i\left(t\right)}$
\ENDFOR
\FOR {$t = K+1,...,$}
\STATE $\textbf{Detect}: \text{determine } \{\hat{\theta}_{i}\} \text{ by Eq.~(\ref{theta_choose})} $
\STATE $\textbf{Balance}: {i\left(t\right) \in \argmax_{i\in\left[K\right]} \mu\left(N_{i}+1 ; \hat{\theta}_{i}\right) }$
\STATE $\textbf{Update}: N_{i\left(t\right)} \leftarrow N_{i\left(t\right)}+1$
\ENDFOR
\end{algorithmic}
\end{algorithm}
\end{minipage}
&
\begin{minipage}{.44\textwidth}
\centering
\begin{algorithm}[H]
\caption{D-\CTO$_{\textnormal{UCB}}$} \label{bcto_alg}
\begin{algorithmic}
\STATE $\textbf{Input}: K, \Theta, \delta$
\STATE $\textbf{Initialization}: N_{i}=0, \text{ } \forall i \in \left[K\right]$
\FOR {$t = 1,2,..,K\times m^{*}_{\text{diff}}\left(\delta/K\right)$}
\STATE $\textbf{Explore}:$
\STATE ${i\left(t\right) \text{ by Round Robin, update } N_{i\left(t\right)}}$
\ENDFOR
\STATE $\textbf{Detect}: \text{determine } \{\hat{\theta}_{i}\} \text{ by Eq.~(\ref{theta_bias_choose})}$
\FOR {$t = K\times m^{*}_{\text{diff}}\left(\delta/K\right)+1,...,$}
\STATE $\textbf{UCB}: i\left(t\right) \text{ according to Eq.~(\ref{ucb_arm})}$
\STATE $\textbf{Update}: N_{i\left(t\right)} \leftarrow N_{i\left(t\right)}+1$
\ENDFOR
\end{algorithmic}
\end{algorithm}
\end{minipage}
\end{tabular}
\end{table}

A few notes on the result: Instead of calculating $m^{*}_{\textnormal{diff}}\left(\delta/K\right)$, it is possible to use any upper bound (e.g., as shown in Appendix E, ${\max_{\theta_{1} \neq \theta_{2} \in \Theta^{2}}  Ddet^{\star \downarrow}_{\theta_{1}, \theta_{2}}\left(\frac{1}{8}  \ln^{-1}\left(\frac{2K}{\delta}\right)\right) }$ rounded to higher even number). We cannot hope for a better rate than $\ln T$ as stochastic MAB is a special case of the \RB\ problem. Finally, we can convert the \textnormal{D-\CTO}$_{\textnormal{UCB}}$ algorithm to have a decision point in each step: at each time step, determine the rotting models according to proximity rule~(\ref{theta_bias_choose}), followed by pulling an arm according to Eq.~(\ref{ucb_arm}). We term this version \textnormal{D-\CTO}$_{\textnormal{SIM-UCB}}$.

\section{Simulations}
\label{sec_sim}
We next compare the performance of the SWA and CTO approaches with benchmark algorithms.

\textbf{Setups} for all the simulations we use Normal distributions with ${\sigma^{2}=0.2}$, and ${T=30,000}$. \\
\textit{Non-Parametric:} ${K=2}$. As for the expected rewards: ${\mu_{1}\left(n\right) = 0.5, \forall n}$, and ${\mu_{2}\left(n\right)=1}$ for its first $7,500$ pulls and $0.4$ afterwards. This setup is aimed to show the importance of not relying on the whole past rewards in the \RB\ setting.\\
\textit{Parametric \AV\ \& \ANV:} ${K=10}$. The rotting models are of the form ${\mu\left(j ; \theta\right) =\left(\textbf{int}\left(\frac{j}{100}\right)+1\right)^{-\theta}}$, where \textbf{int}$\left(\cdot\right)$ is the lower rounded integer, and ${\Theta = \{0.1,0.15,..,0.4\}}$ (i.e., plateaus of length $100$, with decay between plateaus according to $\theta$). ${\{\theta_{i}^{*}\}_{i=1}^{K}}$ were sampled with replacement from $\Theta$, independently across arms and trajectories. ${\{\mu_{i}^{c}\}_{i=1}^{K}}$ (\ANV) were sampled randomly from ${\left[0,0.5\right]^{K}}$.

\textbf{Algorithms} we implemented standard benchmark algorithms for non-stationary \MAB: UCB1 by \citet{FiniteTimeAnalysisMAB}, Discounted UCB (DUCB) and Sliding-Window UCB (SWUCB) by \citet{UpperConfBoundNonStat}. We implemented \CTO$_{\textrm{SIM}}$, D-\CTO$_{\textrm{SIM-UCB}}$, and \wSWA\ for the relevant setups. We note that adversarial benchmark algorithms are not relevant in this case, as the rewards are unbounded.

\textbf{Grid Searches} were performed to determine the algorithms' parameters. For DUCB, following \citet{DiscountedUCB}, the discount factor was chosen from ${\gamma \in \{0.9,0.99,..,0.999999\}}$, the window size for SWUCB from ${\tau \in \{1e3,2e3,..,20e3\}}$, and $\alpha$ for \wSWA\ from ${\{0.2,0.4,..,1\}}$.

\textbf{Performance} for each of the cases, we present a plot of the average regret over $100$ trajectories, specify the number of `wins' of each algorithm over the others, and report the p-value of a paired T-test between the (end of trajectories) regrets of each pair of algorithms. For each trajectory and two algorithms, the \textit{`winner'} is defined as the algorithm with the lesser regret at the end of the horizon.

\begin{figure*}[t]
\vskip 0.05in
%\centering
\begin{center}
\begin{minipage}{.32\textwidth}
  \centering
  \includegraphics[width=0.99\textwidth]{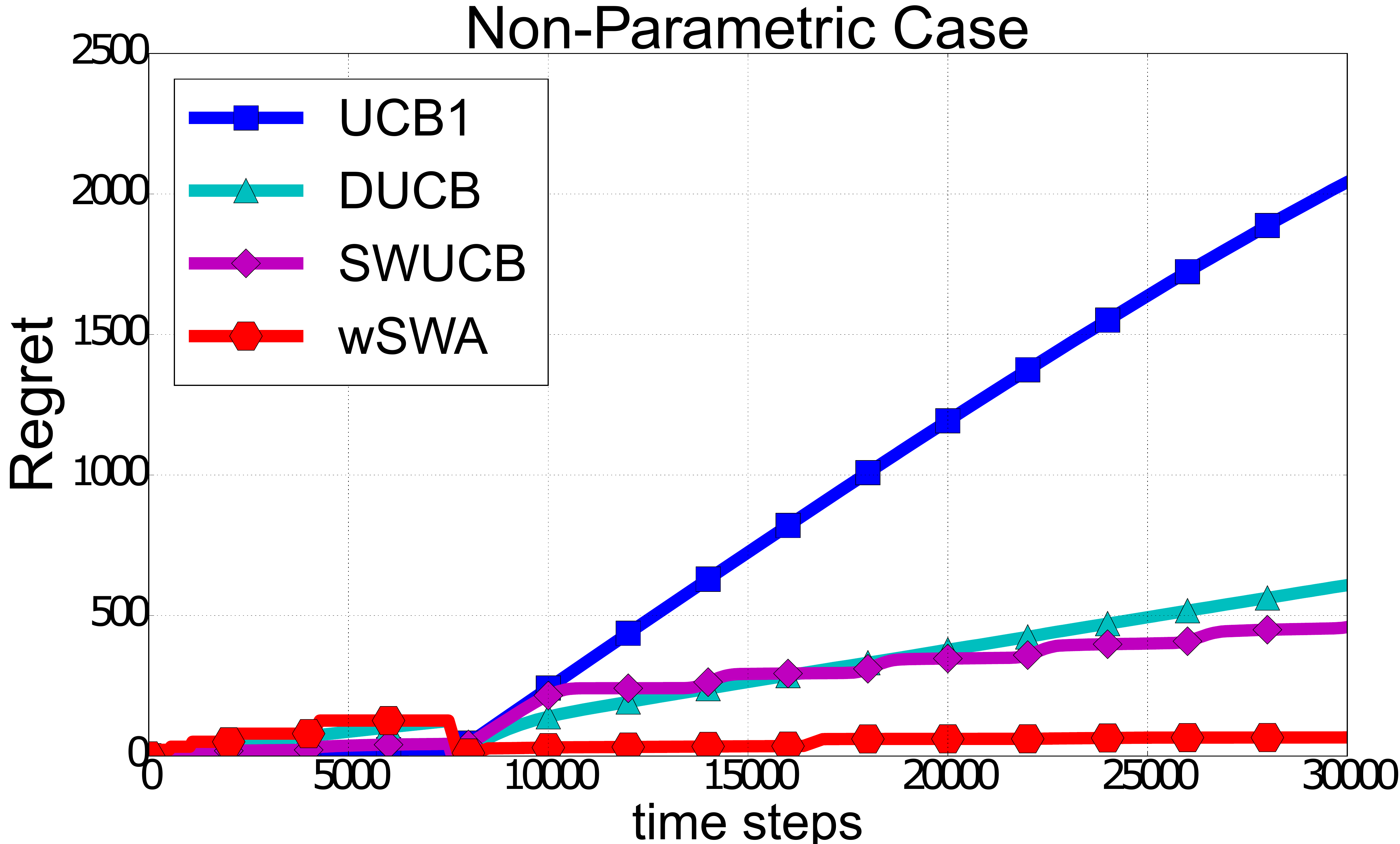}
  %\caption{AV Case: Averaged regret over $100$ trajectories}
  %\label{fig_non_biased_regret}
\end{minipage} \hfill
%\hspace{.05\linewidth}
\begin{minipage}{.32\textwidth}
  \centering
  \includegraphics[width=0.99\textwidth]{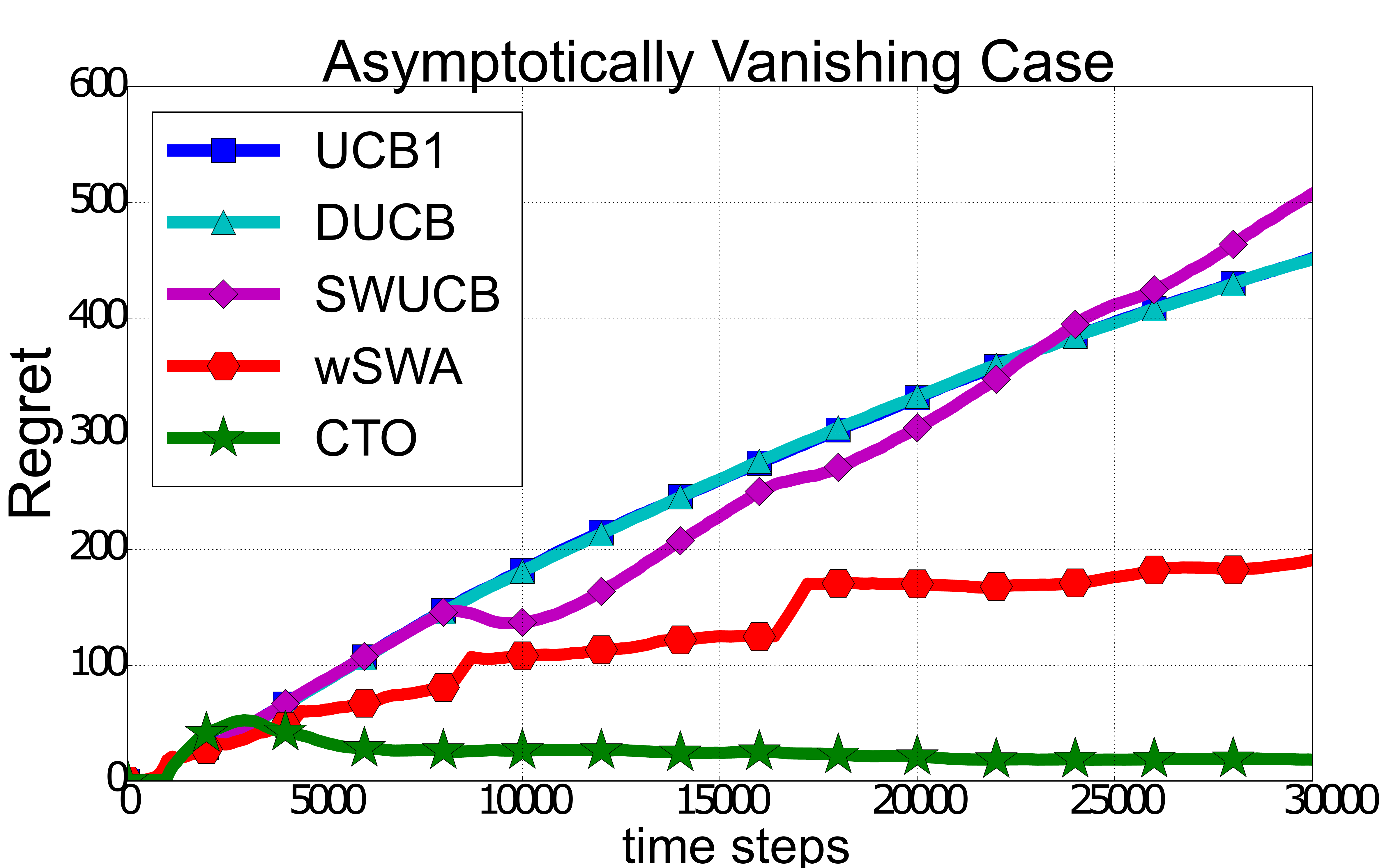}
  %\caption{ANV Case: Averaged regret over $100$ trajectories}
  %\label{fig_biased_regret}
\end{minipage}
\begin{minipage}{.32\textwidth}
  \centering
  \includegraphics[width=0.99\textwidth]{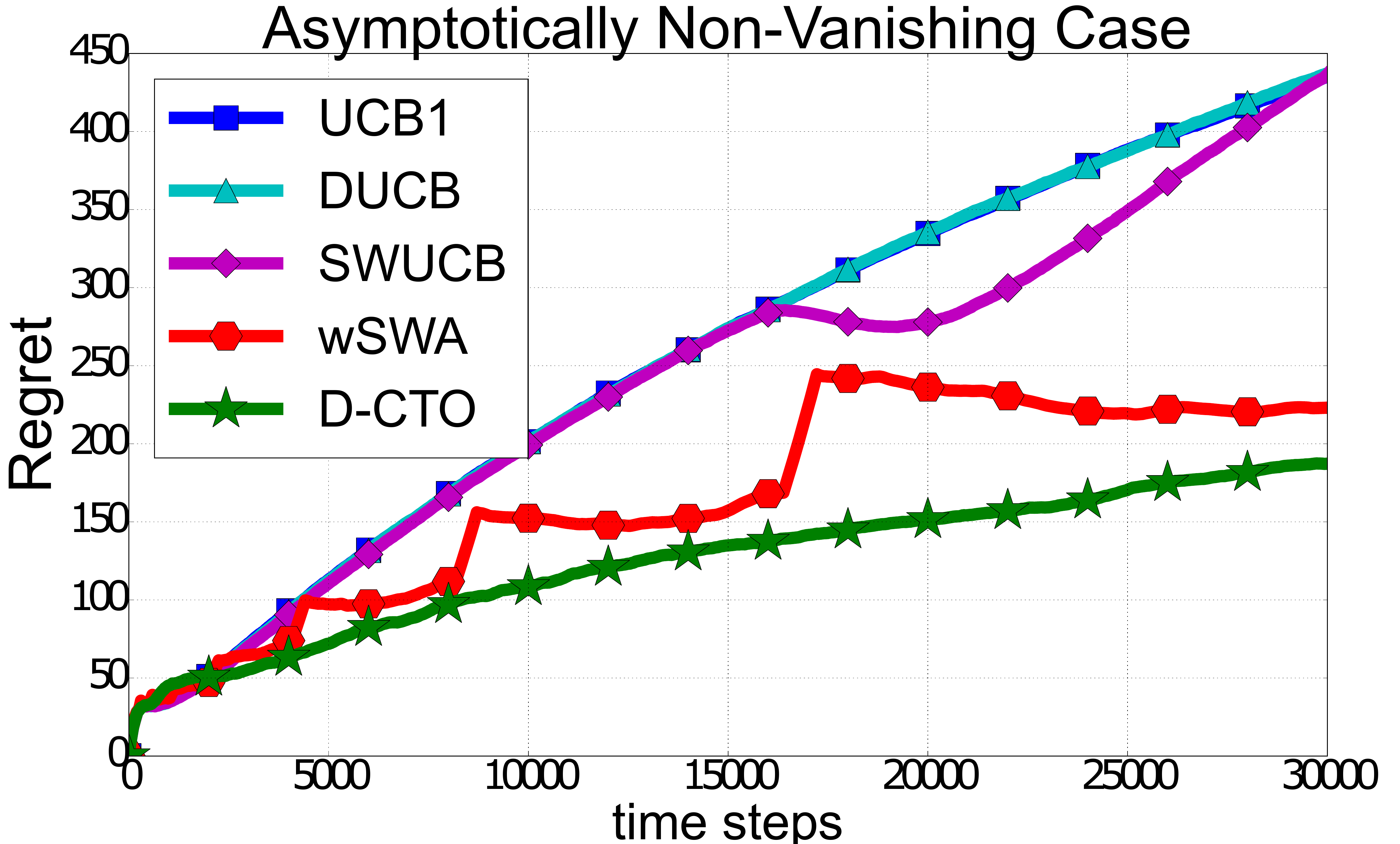}
  %\caption{ANV Case: Averaged regret over $100$ trajectories}
  %\label{qq}
\end{minipage}
\caption{Average regret. Left: non-parametric. Middle: parametric \AV. Right: parametric \ANV \label{fig_sim}}
\end{center}
\vskip -0.2in
\end{figure*}

\textbf{Results} the parameters that were chosen by the grid search are as follows: ${\gamma = 0.999}$ for the non-parametric case, and ${0.999999}$ for the parametric cases. ${\tau = 4e3}$, ${8e3}$, and ${16e3}$ for the non-parametric, \AV, and \ANV\ cases, respectively. ${\alpha=0.2}$ was chosen for all cases. \\
The average regret for the different algorithms is given by Figure~\ref{fig_sim}. Table~\ref{table_results} shows the number of `wins' and p-values. The table is to be read as the following: the entries under the diagonal are the number of times the algorithms from the left column `won' against the algorithms from the top row, and the entries above the diagonal are the p-values between the two. \\
While there is no clear `winner' between the three benchmark algorithms across the different cases, \wSWA, which does not require any prior knowledge, consistently and significantly outperformed them. In addition, when prior knowledge was available and \CTO$_{\textnormal{SIM}}$ or D-CTO$_{\textnormal{UCB-SIM}}$ could be deployed, they outperformed all the others, including \wSWA.

\begin{table*}[t]
\caption{Number of `wins' and p-values between the different algorithms}
\vskip 0.05in
\begin{center}
\begin{small}
\begin{tabular}{>{\centering\arraybackslash}m{0.01\textwidth} | p{0.09\textwidth} || >{\centering\arraybackslash}m{0.12\textwidth} | >{\centering\arraybackslash}m{0.12\textwidth} | >{\centering\arraybackslash}m{0.12\textwidth} | >{\centering\arraybackslash}m{0.12\textwidth} | >{\centering\arraybackslash}m{0.12\textwidth} |}
\multicolumn{2}{c |}{}  & UCB1 & DUCB & SWUCB & wSWA & (D-)CTO \\ \hline \hline
& UCB1 & \cellcolor{gray} & $<$1e-5 & $<$1e-5 & $<$1e-5& \cellcolor{gray} \\ \cline{2-7}
& DUCB & 100 & \cellcolor{gray} & $<$1e-5 & $<$1e-5 & \cellcolor{gray} \\ \cline{2-7}
& SWUCB & 100 & 100 & \cellcolor{gray} & $<$1e-5 &\cellcolor{gray} \\ \cline{2-7}
\multirow{-4}{*}{\begin{sideways}\textbf{NP}\end{sideways}} & wSWA & \textbf{100} & \textbf{100} & \textbf{100} & \cellcolor{gray} & \cellcolor{gray} \\ \hline \hline
& UCB1 & \cellcolor{gray} & 0.81 & $<$1e-5 & $<$1e-5 & $<$1e-5 \\ \cline{2-7}
& DUCB & 55 & \cellcolor{gray} & $<$1e-5 & $<$1e-5 & $<$1e-5 \\ \cline{2-7}
& SWUCB & 15 & 22 & \cellcolor{gray} & $<$1e-5 & $<$1e-5 \\ \cline{2-7}
& wSWA & \textbf{98} & \textbf{99} & \textbf{100} & \cellcolor{gray} & $<$1e-5 \\ \cline{2-7}
\multirow{-5}{*}{\begin{sideways}\textbf{AV}\end{sideways}} & CTO & \textbf{100} & \textbf{100} & \textbf{100} & \textbf{100} & \cellcolor{gray} \\ \hline \hline
& UCB1 & \cellcolor{gray} & 0.54 & 0.83 & $<$1e-5 & $<$1e-5 \\ \cline{2-7}
& DUCB & 40 & \cellcolor{gray} & 0.91 & $<1$e-5 & $<$1e-5 \\ \cline{2-7}
& SWUCB & 50 & 50 & \cellcolor{gray} & $<$1e-5 & $<$1e-5 \\ \cline{2-7}
& wSWA & \textbf{97} & \textbf{98} & \textbf{97} & \cellcolor{gray} & $<$1e-5 \\ \cline{2-7}
\multirow{-5}{*}{\begin{sideways}\textbf{ANV}\end{sideways}} & D-CTO & \textbf{100} & \textbf{100} & \textbf{100} & \textbf{66} & \cellcolor{gray} \\ \hline \hline \end{tabular}
\label{table_results}
\end{small}
\end{center}
\vskip -0.2in
\end{table*}

\section{Related Work}
\label{sec_Rel_work}
%% review different related framworks
% explain differences from our setting
We turn to reviewing related work while emphasizing the differences from our problem.

%% stochastic (stationary)
% (e.g., lai&robbins, auer et al, analysis of TS)
\textbf{Stochastic MAB} In the stochastic MAB setting \citep{AsymEfficient}, the underlying reward distributions are stationary over time. The notion of regret is the same as in our work, but the optimal policy in this setting is one that pulls a fixed arm throughout the trajectory. The two most common approaches for this problem are: constructing Upper Confidence Bounds which stem from the seminal work by \citet{gittins1979bandit} in which he proved that index policies that compute upper confidence bounds on the expected rewards of the arms are optimal in this case (e.g., see \citet{FiniteTimeAnalysisMAB, KLUCB, MABkldiv}), and Bayesian heuristics such as Thompson Sampling which was first presented by \citet{ThompsonLikelihood} in the context of drug treatments (e.g., see \citet{TSfinite, TSfurther, TScomplex}).

%% adversarial
% (e.g., non-stochastic mab problem, prediction learning and games)
\textbf{Adversarial MAB} In the Adversarial MAB setting (also referred to as the Experts Problem, see the book of \citet{PredictionsLearningGames} for a review), the sequence of rewards are selected by an adversary (i.e., can be arbitrary). In this setting the notion of \textit{adversarial regret} is adopted \citep{NonStochasticMAB, BetterAlgs}, where the regret is measured against the best possible fixed action that could have been taken in hindsight. This is as opposed to the \textit{policy regret} we adopt, where the regret is measured against the best sequence of actions in hindsight.

%% hybrid setting (between stocahstic and adversarial)
% (e.g., shie's work, moulines, mortal bandits)
\textbf{Hybrid models} Some past work consider settings between the Stochastic and the Adversarial settings. \citet{UpperConfBoundNonStat} consider the case where the reward distributions remain constant over epochs and change arbitrarily at unknown time instants, similarly to \citet{PiecewiseStatMAB} who consider the same setting, only with the availability of side observations. \citet{MortalMAB} consider the case where arms can expire and be replaced with new arms with arbitrary expected reward, but as long as an arm does not expire its statistics remain the same.

%% non-stationary
% starting from gittins -> rested & restless (e.g., gittins, restless bandits, online learning of rested and restless bandits)
% restless - fundamentally different (e.g., chinese person work, brownian motion, zeevi's work)
% rested (e.g., mab with known trend -> not decreasing, only gain is missing ; tight policy regret bounds for.. -> non-stochastic)
\textbf{Non-Stationary MAB} Most related to our problem is the so-called Non-Stationary MAB. Originally proposed by \citet{DynamicAllocationIndex}, who considered a case where the reward distribution of a chosen arm can change, and gave rise to a sequence of works (e.g., \citet{ArmAcquiring, OnlineLearningRestedRestless}) which were termed \textit{Restless Bandits} and \textit{Rested Bandits}. In the \textit{Restless Bandits} setting, termed by \citet{RestlessBanditsAllocation}, the reward distributions change in each step according to a known stochastic process. \citet{TimeDecayingBandits} consider the case where each arm decays according to a linear combination of decaying basis functions. This is similar to our parametric case in that the reward distributions decay according to possible models, but differs fundamentally in that it belongs to the \textit{Restless Bandits} setup (ours to the \textit{Rested Bandits}). More examples in this line of work are \citet{BrownianRestless} who consider evolution of rewards according to Brownian motion, and \citet{StocMABNonStat} who consider bounded total variation of expected rewards. The latter is related to our setting by considering the case where the total variation is bounded by a constant, but significantly differs by that it considers the case where the (unknown) expected rewards sequences are not affected by actions taken, and in addition requires bounded support as it uses the EXP3 as a sub-routine. In the \textit{Rested Bandits} setting, only the reward distribution of a chosen arm changes, which is the case we consider. An optimal control policy (reward processes are known, no learning required) to bandits with non-increasing rewards and discount factor was previously presented (e.g., \citet{mandelbaum1987continuous}, and \citet{kaspi1998multi}). \citet{TightPolicyRegret} (2016) consider the case where the reward decays (as we do), but with no statistical noise (deterministic rewards), which significantly simplifies the problem. Another somewhat closely related setting is suggested by \citet{MABKnownTrend}, in which statistical noise exists, but the expected reward shape is known up to a multiplicative factor.

\section{Discussion}
\label{Sec_dis}
We introduced a novel variant of the \textit{Rested Bandits} framework, which we termed \textit{Rotting Bandits}. This setting deals with the case where the expected rewards generated by an arm decay (or generally do not increase) as a function of pulls of that arm. This is motivated by many real-world scenarios.

We first tackled the non-parametric case, where there is no prior knowledge on the nature of the decay. We introduced an easy-to-follow algorithm accompanied by theoretical guarantees.

We then tackled the parametric case, and differentiated between two scenarios: expected rewards decay to zero (\AV), and decay to different constants (\ANV). For both scenarios we introduced suitable algorithms with stronger guarantees than for the non-parametric case: For the \AV\ scenario we introduced an algorithm for ensuring, in expectation, regret upper bounded by a term that decays to zero with the horizon. For the \ANV\ scenario we introduced an algorithm for ensuring, with high probability, regret upper bounded by a horizon-dependent rate which is optimal for the stationary case.

We concluded with simulations that demonstrated our algorithms' superiority over benchmark algorithms for non-stationary MAB. We note that since the \RB\ setting is novel, there are not suitable available benchmarks, and so this paper also serves as a benchmark.

For future work we see two main interesting directions: (i) show a lower bound on the regret for the non-parametric case, and (ii) extend the scope of the parametric case to continuous parameterization.

\paragraph{Acknowledgment}
The research leading to these results has received funding from the European Research Council under the European Union's Seventh Framework Program (FP/2007-2013) / ERC Grant Agreement n. 306638

\begin{small}
%\medskip
\bibliographystyle{abbrvnat}
\bibliography{rotting_bandits_ref}
\end{small}

\newpage
\appendix

\section{Hoeffding's Inequality for Sub-Gaussian RVs}
Let $X_{1},..,X_{n}$ be independent, mean-zero, $\sigma_{i}^{2}$-sub-Gaussian random variables. Then for all $t \geq 0$,
\begin{equation} 
\prob{\sum_{i=1}^{n} X_{i} \geq t} \leq \exp \bigg\{-\frac{t^{2}}{2\sum_{i=1}^{n}\sigma_{i}^{2}}\bigg\}
\label{hoeff_ineq}
\end{equation}

\section{Optimal Policy}
\subsection{Proof of Lemma~\ref{lemma_opt_policy}}
In this section we show that $\pi^{\textrm{max}}$, defined by Eq.~(\ref{opt_policy}) is an optimal policy for the \RB\ problem. \\
Assume on the contrary, that $\pi^{\textrm{max}}$ is not an optimal policy. Thus, there exists a time horizon, $T$, for which there exists some other policy $\pi^{\textrm{cand}}$ that satisfies ${J\left(T ; \pi^{\textrm{cand}}\right) > J\left(T ; \pi^{\textrm{max}}\right)}$. \\
Let $m$ be the first time step in which $\pi^{\textrm{cand}}$ deviates from $\pi^{\textrm{max}}$, since ${J\left(T ; \pi^{\textrm{cand}}\right) > J\left(T ; \pi^{\textrm{max}}\right)}$ we infer that $m \leq T$ (i.e., there is such time step). Let $\tilde{\pi}$ be a policy defined by,
\begin{equation} \nonumber
\tilde{\pi}\left(t\right) = \begin{cases}
	\pi^{\textrm{cand}}\left(t\right), & \text{if } t<m \\
	\argmax_{i\in\left[K\right]} \{\mu\left(N_{i}\left(m\right)+1 ;  \theta^{*}_{i}\right)\}, & \text{if } t=m \\
	\pi^{\textrm{cand}}\left(t-1\right), & \text{if } t>m
	\end{cases}
\end{equation}
where if there exist more than one member in ${\argmax_{i \in \left[K\right]}\{\mu\left(N_{i}\left(m\right)+1 ; \theta_{i}^{*}\right)\}}$, $\tilde{\pi}$ chooses the same action as $\pi^{\textrm{max}}$. That is, $\tilde{\pi}$ mimics $\pi^{\textrm{cand}}$ until time step $m$, then plays according to $\argmax$ rule, and then re-mimics $\pi^{\textrm{cand}}$. Let $\mu_{m}$, $\mu_{T}$ be the expected rewards of the arms that $\tilde{\pi}$ chose at the $m^{th}$ time step, and that $\pi^{\textrm{cand}}$ chose at the $T^{th}$ time step, respectively. It is easy to see that,
\begin{equation} 
J\left(T ; \tilde{\pi}\right) - J\left(T ; \pi^{cand}\right) = \mu_{m} - \mu_{T} \geq 0
\end{equation}
where the second transition holds by the $\argmax$ rule combined with the assumption that the expected rewards are non-increasing (assumption~\ref{assm_rotting}). Thus, $J\left(T ; \tilde{\pi}\right) \geq J\left(T ; \pi^{\textrm{cand}}\right)$. If we apply the above logic steps recursively, we obtain a series of policies with non-decreasing values of expected total reward $J\left(T ; \cdot\right)$, where the series ends when there is no time step which deviates from $\pi^{\textrm{max}}$, i.e., $J\left(T ; \pi^{\textrm{max}}\right) \geq J\left(T ; \pi^{\textrm{cand}}\right)$, in contradiction to $\pi^{\textrm{max}}$ being non-optimal. Thus, we infer that $\pi^{\textrm{max}}$ is indeed an optimal policy.

\section{Non-Parametric Case}
\subsection{Proof of Thm.~\ref{thm_swa_known_horizon}}
We define,
\begin{equation} \nonumber
\begin{cases}
M & = \ceil{\alpha 4^{2/3} \sigma^{2/3} K^{-2/3} T^{2/3} \ln^{1/3} \left(\sqrt{2} T\right)} \\
q & = \alpha^{-1/2} 2^{1/3} \sigma^{2/3} K^{1/3} T^{-1/3} \ln^{1/3} \left(\sqrt{2} T\right)
\end{cases}
\end{equation}
and start by making two useful observations: 

\underline{Observation 1:} By Hoeffding's Inequality we have,
\begin{equation}
P\left(| \bar{X}_{M} - \mathbb{E}\left[\bar{X}_{M}\right]| \geq q \right) \leq \frac{1}{T^{2}}
\end{equation}
where $\bar{X}_{M}$ is the empirical average of $M$ independent $\sigma^{2}$ sub-Gaussian samples. 

\underline{Observation 2:} Since the expected rewards of an arm only depends on the time it is being pulled (and not on the time step itself), the expected total reward of a policy only depends on the number of pulls of the different arms (and not on the order of pulls).

From now on we assume that ${|\bar{X}_{M}-\mathbb{E}\left[\bar{X}_{M}\right]| < q}$ (see Observation~1) for all arms throughout the trajectory, and later address the case where it is violated. 

\newpage
\textit{\textbf{Step 1:}} bound the number of significantly sub-optimal pulls.

In what is following we prove by induction that for all the ends of time steps $t \in \left[T\right]$, by applying \SWA, there is no arm $j$ for which,
\begin{equation}
\bigg\{|n| :  \mu_{j}\left(N_{j}^{\pi^{\textrm{\SWA}}}\left(t\right)-n\right) < \max_{i \in \left[K\right]} \left[\mu_{i}\left(N_{i}^{\pi^{\textrm{\SWA}}}\left(t\right)\right)\right] - 2q, \quad n \in \mathbb{N}_{0} \bigg\} > M
\label{eq_induction}
\end{equation}
where $N_{i}^{\pi^{\textrm{\SWA}}}\left(t\right)$ is the number of pulls of arms $i$ at time $t$ induced by policy $\pi^{\textrm{\SWA}}$, which is defined by the \SWA\ algorithm. That is, following \SWA\ ensures that for all time steps, no arm would be pulled more than $M$ times in which its expected reward is at least $2q$ lower than the expected reward of the (current) optimal arm.\\
\textit{Basis:} for all the ends of time steps $t \in \{1,..,KM\}$ this holds trivially since, by the definition of \SWA\ we pull each arm exactly $M$ times. \\
\textit{Inductive hypothesis:} Assume that the above statement holds for the end of time step $t'$ such that, $KM \leq t' < T$. \\
\textit{Inductive step:} We show that the above statement holds for the end of time step $t'+1$. By the non-increasing Assumption~\ref{assm_rotting} we note two things: (1) The RHS of the inner inequality in Eq.~(\ref{eq_induction}) is non-increasing in $t$, thus if the inequality did not hold for some arm $j$ at the end of time step $t'$ it can only hold for it at the end of $t'+1$ if \SWA\ pulls arm $j$ in that round. (2) The number of $n$s for which the inequality holds for some arm $j$ can increase only by one at each time step. Combining the two with our inductive hypothesis we simply need to show that if for some arm $j$, Eq.~(\ref{eq_induction}) holds with equality (i.e., the number of $n$s is $M$), that arm would not be pulled in $t'+1$. By the non-increasing Assumption~\ref{assm_rotting} we know that the last $M$ expected rewards of arm $j$ are those who are at least $2q$ lower. Let ${i^{*} \in \argmax_{i \in \left[K\right]} \left[\mu_{i}\left(N_{i}^{\pi^{\textrm{\SWA}}}\left(t'+1\right)\right)\right]}$ (if this set contains more than one arm, choose arbitrarily). We have,
\begin{multline}
\frac{1}{M} \sum_{n=N_{j}^{\pi^{\textrm{\SWA}}}\left(t'+1\right)-M+1}^{N_{j}^{\pi^{\textrm{\SWA}}}\left(t'+1\right)} r_{j}^{n} \overset{(1)}{<} \mathbb{E}\left[\frac{1}{M} \sum_{n=N_{j}^{\pi^{\textrm{\SWA}}}\left(t'+1\right)-M+1}^{N_{j}^{\pi^{\textrm{\SWA}}}\left(t'+1\right)} r_{j}^{n}\right] + q \overset{(2)}{\leq} \\
\mu_{j}\left(N_{j}^{\pi^{\textrm{\SWA}}}\left(t'+1\right)-M+1\right) + q  \overset{(3)}{<} \mu_{i^{*}}\left(N_{i^{*}}^{\pi^{\textrm{\SWA}}}\left(t'+1\right)\right) - q \overset{(4)}{\leq} \\
\mathbb{E}\left[\frac{1}{M} \sum_{n=N_{i^{*}}^{\pi^{\textrm{\SWA}}}\left(t'+1\right)-M+1}^{N_{i^{*}}^{\pi^{\textrm{\SWA}}}\left(t'+1\right)} r_{i^{*}}^{n}\right] - q \overset{(5)}{<} \frac{1}{M} \sum_{n=N_{i^{*}}^{\pi^{\textrm{\SWA}}}\left(t'+1\right)-M+1}^{N_{i^{*}}^{\pi^{\textrm{\SWA}}}\left(t'+1\right)} r_{i^{*}}^{n}
\end{multline}
where $(1)$ and $(5)$ hold by our assumption regarding ${|\bar{X}_{M}-\mathbb{E}\left[\bar{X}_{M}\right]| < q}$, $(2)$ and $(4)$ hold by the non-increasing Assumption~\ref{assm_rotting}, and $(3)$ holds by the definition of the inequality in Eq.~(\ref{eq_induction}). Since the \SWA\ algorithm chooses in the Balance step according to the empirical averages of the last $M$-pulls of each arm, we infer that arm $j$ would not be pulled ($i^{*}$ has higher empirical average). This concludes the inductive step proof, and hence our statement holds.

\textit{\textbf{Step 2:}} bound ${J\left(T ; \pi^{\widehat{\textrm{max}}}\right) - J\left(T ; \pi^{\textrm{SWA}}\right)}$.

Let $\pi^{\widehat{\textrm{max}}}$ be a policy defined by,
\begin{equation}
\pi^{\widehat{\textrm{max}}}\left(t\right) \in \argmax_{i \in \left[K\right]} \{ \mu_{i}\left(N_{i}\left(t\right)\right) \}
\end{equation}
where we first pull each arm once using Round-Robin (before following the above rule), and in a case of tie, break it using the smallest index. \\
Define
\begin{equation}
I^{\widehat{\textrm{max}}}\left(T\right) = \bigg\{\left(i_{t}^{\widehat{\textrm{max}}},n_{t}^{\widehat{\textrm{max}}}\right)\bigg\}_{t=1}^{T}
\end{equation}
to be the (deterministic) set of tuples induced by applying $\pi^{\widehat{\textrm{max}}}$, where ${i_{t}^{\widehat{\textrm{max}}}}$ is the arm chosen at time step $t$, and $n_{t}^{\widehat{\textrm{max}}}$ is the time it is being pulled. In the same manner, we define the (stochastic) set ${I^{\textrm{\SWA}}\left(T\right)}$, composed of ${\left( i_{t}^{\textrm{\SWA}}, n_{t}^{\textrm{\SWA}} \right)}$ tuples, which induced by applying ${\pi^{\textrm{\SWA}}}$. We further define ${I^{\widehat{\textrm{max}}}_{\backslash \textrm{\SWA}}\left(T\right) = I^{\widehat{\textrm{max}}}\left(T\right) \backslash \{I^{\widehat{\textrm{max}}}\left(T\right) \cap I^{\textrm{\SWA}}\left(T\right) \}  }$, and ${I^{\textrm{\SWA}}_{\backslash \widehat{\textrm{max}}}\left(T\right) = I^{\textrm{\SWA}}\left(T\right) \backslash \{I^{\widehat{\textrm{max}}}\left(T\right) \cap I^{\textrm{\SWA}}\left(T\right) \}  }$, and also ${\mu^{\textrm{\SWA}}_{\textrm{max}}\left(T+1\right) = \max_{i \in \left[K\right]} \left[ \mu_{i}\left(N_{i}^{\pi^{\textrm{\SWA}}}\left(T+1\right)\right) \right] }$. By Observation~2, the difference in the policies expected total rewards only depends on these number of pull sets. Since both policies start with one Round-Robin pulls of the arms we have,
\begin{equation}
\begin{split}
J\left(T ; \pi^{\widehat{\textrm{max}}}\right) - J\left(T ; \pi^{\textrm{SWA}}\right) & = \sum_{\mathclap{\left(i_{t}^{\widehat{\textrm{max}}}, n_{t}^{\widehat{\textrm{max}}}\right) \in I^{\widehat{\textrm{max}}}}} \mu_{i_{t}^{\widehat{\textrm{max}}}}\left(n_{t}^{\widehat{\textrm{max}}}\right) - \sum_{\mathclap{\left(i_{t}^{\textrm{\SWA}}, n_{t}^{\textrm{\SWA}}\right) \in I^{\textrm{\SWA}}}} \mu_{i_{t}^{\textrm{\SWA}}}\left( n_{t}^{\textrm{\SWA}} \right) \\
& = \sum_{\mathclap{\left(i_{t}^{\widehat{\textrm{max}}}, n_{t}^{\widehat{\textrm{max}}}\right) \in I^{\widehat{\textrm{max}}}_{\backslash \textrm{\SWA}}}} \mu_{i_{t}^{\widehat{\textrm{max}}}}\left(n_{t}^{\widehat{\textrm{max}}}\right) - \sum_{\mathclap{\left(i_{t}^{\textrm{\SWA}}, n_{t}^{\textrm{\SWA}}\right) \in I^{\textrm{\SWA}}_{\backslash \widehat{\textrm{max}}}}} \mu_{i_{t}^{\textrm{\SWA}}}\left( n_{t}^{\textrm{\SWA}} \right) \\
& \leq \mu_{\textrm{max}}^{\textrm{\SWA}}\left(T+1\right) \times |I^{\widehat{\textrm{max}}}_{\backslash \textrm{\SWA}}| - 0 \times KM \\
& \qquad - \left(\mu_{\textrm{max}}^{\textrm{\SWA}}\left(T+1\right) - 2q\right) \times \left( |I^{\widehat{\textrm{max}}}_{\backslash \textrm{\SWA}}| - KM \right) \\
& \leq KM \max_{i \in \left[K\right]} \mu_{i}\left(1\right) + 2qT
\end{split}
\end{equation}
The first inequality holds by: (1) the non-increasing Assumption~\ref{assm_rotting} implies that all the tuples in ${I^{\widehat{\textrm{max}}}_{\backslash \textrm{\SWA}}}$ correspond to expected reward upper bounded by ${\mu_{\textrm{max}}^{\textrm{\SWA}}\left(T+1\right)}$, and (2) by what we showed in \textit{Step 1}, there are at most $KM$ members in ${I^{\textrm{\SWA}}_{\backslash \widehat{\textrm{max}}}}$ that are more than $2q$ below ${\mu_{\textrm{max}}^{\textrm{\SWA}}\left(T+1\right)}$, and the positiveness of the expected rewards by Assumption~\ref{assm_rotting}. The second inequality holds by trivially bounding ${\mu_{\textrm{max}}^{\textrm{\SWA}}\left(T+1\right) \leq \max_{i \in \left[K\right]} \mu_{i}\left(1\right)}$, and ${|I^{\widehat{\textrm{max}}}_{\backslash \textrm{\SWA}}| =  |I^{\textrm{\SWA}}_{\backslash \widehat{\textrm{max}}}| \leq T}$. 

Finally, we note that all the above analysis was done assuming that ${|\bar{X}_{M}-\mathbb{E}\left[\bar{X}_{M}\right]| < q}$ for all arms throughout the trajectory, and we now address the case where it is violated. By Observation~1, the probability of the inequality to be violated ${\leq 1/T^{2}}$. The number of times this inequality is tested throughout the trajectory is bounded by $KT$ (for each of the arms, in every time step, during the Balance step), and if the inequality is violated (even once) then ${J\left(T ; \pi^{\widehat{\textrm{max}}}\right) - J\left(T ; \pi^{\textrm{SWA}}\right)}$ is trivially bounded by ${T \max_{i \in \left[K\right]} \mu_{i}\left(1\right)}$ according to the non-increasing Assumption~\ref{assm_rotting}. Thus, we infer that in expectation we have,
\begin{equation}
J\left(T ; \pi^{\widehat{\textrm{max}}}\right) - J\left(T ; \pi^{\textrm{SWA}}\right) \leq KM \max_{i \in \left[K\right]} \mu_{i}\left(1\right) + 2qT + K \max_{i \in \left[K\right]} \mu_{i}\left(1\right)
\end{equation}

\textit{\textbf{Step 3:}} bound the regret.

We bound the regret using our previous obtained result for $\pi^{\widehat{\textrm{max}}}$ by,
\begin{equation}
\begin{split}
\mathcal{R}\left(T ; \pi^{\textrm{\SWA}}\right) & = \max_{\pi \in \Pi} \{ J\left(T ; \pi\right) \} - J\left(T ; \pi^{\textrm{\SWA}}\right) \\ 
& = J\left(T ; \pi^{\textrm{max}}\right) - J\left(T ; \pi^{\textrm{\SWA}}\right) \\
& = \left(J\left(T ; \pi^{\textrm{max}}\right)- J\left(T ; \pi^{\widehat{\textrm{max}}}\right)\right) + \left(J\left(T ; \pi^{\widehat{\textrm{max}}}\right) - J\left(T ; \pi^{\textrm{\SWA}}\right)\right) \\
& \leq K \max_{i \in \left[K\right]} \mu_{i}\left(1\right) + \left(J\left(T ; \pi^{\widehat{\textrm{max}}}\right) - J\left(T ; \pi^{\textrm{\SWA}}\right)\right) \\
& \leq 2K \max_{i \in \left[K\right]} \mu_{i}\left(1\right) +
KM \max_{i \in \left[K\right]} \mu_{i}\left(1\right) + 2qT \\
& = \left(\alpha \max_{i \in \left[K\right]} \mu_{i}\left(1\right) + \alpha^{-1/2}\right) 4^{2/3} \sigma^{2/3} K^{1/3} T^{2/3} \ln ^{1/3} \left(\sqrt{2} T \right) + 3K \max_{i \in \left[K\right]} \mu_{i}\left(1\right)
\end{split}
\end{equation}
where the first equality holds by Lemma~\ref{lemma_opt_policy}, the first inequality holds by Theorem~3 in \citet{TightPolicyRegret}~[2016], the second inequality holds by the bound we found in \textit{Step~2}, and the last equality holds by plugging in the definition for $M$ and $q$. This establishes Theorem~\ref{thm_swa_known_horizon}.

\subsection{Proof of Corollary~\ref{crl_swa_unknown_horizon}}
For convenience, we define the following objects: ${\mathcal{R}\left(t_{1} \rightarrow t_{2} ; \pi \right)}$ is the regret accumulated between time steps $t_{1}$ and $t_{2}$ (included), by applying policy $\pi$ consistently. ${\mathcal{R}\left(t_{1} \rightarrow t_{2} ; \pi_{2} | \pi_{1}\left(t_{1}\right) \right)}$ is the regret accumulated between time steps $t_{1}$ and $t_{2}$, by applying $\pi_{1}$ until time step $t_{1}$, and then $\pi_{2}$ for the measured time steps. We define similar objects for the expected total reward, ${J}$.

We note that,
\begin{equation}
J\left(t_{1} \rightarrow t_{2} ; \pi^{\textrm{max}}\right) \leq J\left(t_{1} \rightarrow t_{2} ; \pi^{\textrm{max}} \bigg| \pi\left(t_{1}\right) \right) , \quad \forall \pi \in \Pi
\label{eq_total_reward_switch_policy}
\end{equation}
The above inequality can be understood by the following argument: consider a decreasing sorted list of all the expected rewards across all arms. By Assumption~\ref{assm_rotting}, at each time step, ${\pi^{\textrm{max}}}$ simply pulls an arm corresponding to the highest element in that list, that was not previously pulled (independently of previous pulls). \\
Thus, ${J\left(t_{1} \rightarrow t_{2} ; \pi^{\textrm{max}}\right)}$ is the sum of the $t_{1}^{\textrm{th}}$ to $t_{2}^{\textrm{th}}$ elements in this list, which is the lowest possible sum of the ${|t_{2}-t_{1}+1|}$ highest elements in the list, following any $|t_{1}-1|$ pulls.

Consider the $n^{\textrm{th}}$ iteration of \wSWA. i.e., between time steps ${t_{1}=2^{n-1}}$ and ${t_{2}=\min \left[2^{n}-1,T\right]}$. We have,
\begin{equation}
\begin{split}
\mathcal{R}\left(t_{1} \rightarrow t_{2} ; \pi^{\textrm{\wSWA}} \right) & \overset{(1)}{=} J\left( t_{1} \rightarrow t_{2} ; \pi^{\textrm{max}} \right) - J\left(t_{1} \rightarrow t_{2} ; \pi^{\textrm{\wSWA}}\right) \\
& \overset{(2)}{=} J\left( t_{1} \rightarrow t_{2} ; \pi^{\textrm{max}} \bigg| \pi^{\textrm{max}}\left(t_{1}\right) \right) - J\left(t_{1} \rightarrow t_{2} ; \pi^{\textrm{\wSWA}} \bigg| \pi^{\textrm{\wSWA}}\left(t_{1}\right) \right) \\
& \overset{(3)}{\leq} J\left( t_{1} \rightarrow t_{2} ; \pi^{\textrm{max}} \bigg| \pi^{\textrm{\wSWA}}\left(t_{1}\right) \right) - J\left(t_{1} \rightarrow t_{2} ; \pi^{\textrm{\wSWA}} \bigg| \pi^{\textrm{\wSWA}}\left(t_{1}\right) \right) \\
& \overset{(4)}{=} J\left( t_{1} \rightarrow t_{2} ; \pi^{\textrm{max}} \bigg| \pi^{\textrm{\wSWA}}\left(t_{1}\right) \right) - J\left(t_{1} \rightarrow t_{2} ; \pi^{\textrm{\SWA}} \bigg| \pi^{\textrm{\wSWA}}\left(t_{1}\right) \right) \\
& \overset{(5)}{=} \mathcal{R}\left(t_{1} \rightarrow t_{2} ; \pi^{\textrm{\SWA}} \bigg| \pi^{\textrm{\wSWA}}\left(t_{1}\right)  \right) \\
& \overset{(6)}{\leq} \mathcal{R}_{\textrm{bound}}\left( t_{2} - t_{1} +1 \right) 
\end{split}
\label{eq_wswa_regret_summand}
\end{equation}
where $(1)$ and $(2)$ hold by definition. $(3)$ holds by Eq.~(\ref{eq_total_reward_switch_policy}). $(4)$ by noting the \wSWA\ applies \SWA\ between $t_{1}$ and $t_{2}$. $(5)$ by definition. $(6)$ by observing that it is the regret of a known horizon problem that holds Assumption~\ref{assm_rotting}, thus we can use the upper bound from Theorem~\ref{thm_swa_known_horizon}, denoted by ${\mathcal{R}_{\textrm{bound}}}$.

Let $\tilde{n} = \floor{\log_{2}T}+1$, thus ${2^{\tilde{n}-1} \leq T \leq 2^{\tilde{n}}-1}$, and we have,
\begin{equation}
\begin{split}
\mathcal{R}\left(T ; \pi^{\textrm{\wSWA}}\right) & \overset{(1)}{=} \sum_{y=1}^{\tilde{n}-1} \mathcal{R}\left( 2^{y-1} \rightarrow 2^{y}-1 ; \pi^{\textrm{\wSWA}} \right) + \mathcal{R}\left(2^{\tilde{n}-1} \rightarrow T ; \pi^{\textrm{\wSWA}}  \right) \\
& \overset{(2)}{\leq} \sum_{y=1}^{\tilde{n}-1} \mathcal{R}_{\textrm{bound}}\left( 2^{y-1} \right) + \mathcal{R}_{bound}\left(T - 2^{\tilde{n}-1} + 1 \right) \\
& \overset{(3)}{\leq} \sum_{y=0}^{\tilde{n}-1} \mathcal{R}_{\textrm{bound}}\left( 2^{y} \right) \\
& \overset{(4)}{=} \sum_{y=0}^{\tilde{n}-1} \left[ A 2^{2y/3} \ln ^{1/3} \left(2^{y+1/2} \right) + B\right] \\
& \overset{(5)}{\leq} A \ln^{1/3} \left(\sqrt{2} T\right) \sum_{y=0}^{\tilde{n}-1} 2^{2y/3} + B \left( \log_{2}T+1 \right)\\
& \overset{(6)}{\leq} A 2^{5/3} T^{2/3} \ln^{1/3} \left(\sqrt{2}T \right)  + B \left( \log_{2}T+1 \right)
\end{split} 
\end{equation}
where $(1)$ holds by dividing the horizon and noting that the regret is additive. $(2)$ holds by Eq~(\ref{eq_wswa_regret_summand}). $(3)$ holds by noting that both Theorem~3 from \citet{TightPolicyRegret} and \textit{Step~1} from the proof of Theorem~\ref{thm_swa_known_horizon} hold for any $t \in \left[T\right]$, thus the upper bound ${\mathcal{R}_{\textrm{bound}}}$ holds for any ${t \in \left[T\right]}$ (clearly, by plugging $T$ in the bound). $(4)$ holds by plugging ${\mathcal{R}_{\textrm{bound}}}$ and defining ${A = \left(\alpha \max_{i \in \left[K\right]} \mu_{i}\left(1\right) + \alpha^{-1/2}\right) 4^{2/3} \sigma^{2/3} K^{1/3}}$, and ${B = 3K \max_{i \in \left[K\right]} \mu_{i}\left(1\right)}$. $(5)$ holds by monotonicity of the logarithm, and noting that $A$ and $B$ are independent of $y$. Finally, $(6)$ holds as a sum of a geometric series, and simple algebra. \\
Plugging back $A$ and $B$, we establish Corollary~\ref{crl_swa_unknown_horizon}.

\section{Parametric Case}
\subsection{Proof of Thm.~\ref{thm_cto_sim_pure}}
\textbf{Bounding number of steps to optimality} \\
We first characterize the bound, and later show feasibility (i.e., that the analysis we show here indeed holds within the horizon).\\
Similar to the definition of ${m^{*}_{\textrm{diff}}\left(p ; \theta_{i}^{*}\right)}$  and ${m^{*}_{\textrm{diff}}\left(p\right)}$, we define ${m^{*}\left(p ; \theta_{i}^{*}\right)}$ as the solution to optimization problem~(\ref{opt_prob_dcto}) using Eq.~(\ref{theta_choose}) as the proximity rule to hypothesize $\hat{\theta}$, and ${m^{*}\left(p\right) = \max_{\theta \in \Theta} m^{*}\left(p ; \theta\right)}$.

Let $T$ be some \textit{unknown} horizon. We first show that ${m^{*}\left(\frac{1}{KT^{2}}\right)}$ is finite. Define,
\begin{equation} 
\theta_{i}^{'}\left(\tilde{m}\right) = \argmin_{\theta \neq \theta_{i}^{*}} \bigg\{  \bigg|\sum_{j=1}^{\tilde{m}}\mu\left(j ; \theta_{i}^{*}\right) - \sum_{j=1}^{\tilde{m}}\mu\left(j ; \theta\right)\bigg|\bigg\}
\end{equation}
Thus we have, when we sample only from arm $i$,
\begin{equation} 
\begin{split}
P\left( \hat{\theta}_{i}\left(\tilde{m}\right) \neq \theta^{*}_{i} \right) & =  P\left( \exists \theta \neq \theta^{*}_{i} : |Y\left(i,\tilde{m}; \theta\right)| \leq |Y\left(i ,\tilde{m}; \theta^{*}_{i}\right)| \right) \\
& \leq P\left( \bigg| \sum_{j=1}^{\tilde{m}}r_{j}^{i} -\sum_{j=1}^{\tilde{m}} \mu\left(j ; \theta_{i}^{*}\right) \bigg| > \frac{1}{2}  \bigg| \sum_{j=1}^{\tilde{m}}\mu\left(j ; \theta_{i}^{*}\right) - \sum_{j=1}^{\tilde{m}}\mu\left(j ; \theta_{i}^{'}\left(\tilde{m}\right)\right) \bigg| \right)  \\
& \leq 2  \exp \bigg\{ - \frac{1}{8 \times det_{\theta^{*}_{i}, \theta_{i}^{'}\left(\tilde{m}\right)}\left(\tilde{m}\right)} \bigg\}
\end{split}
\end{equation}
where the first inequality holds by inclusion of events, and the second inequality holds by Eq.~(\ref{hoeff_ineq}) and the definition of $det_{\theta^{*}_{i},\theta_{i}^{'}}$. \\
Since trivially ${bal\left(n\right) \geq n}$, by assumption~\ref{assm_sim}, there exists a finite $\tilde{m}$, for which,
\begin{equation} 
\max_{\theta_{1} \neq \theta_{2} \in \Theta^{2}} \bigg\{ det_{\theta_{1}, \theta{2}}\left(\tilde{m}\right) \bigg\} \leq \frac{1}{8}  \ln^{-1} \left(2KT^{2}\right)
\end{equation}
Therefore, if we plug $\tilde{m}$ back in to the above equation we get,
\begin{equation} 
2  \exp \bigg\{ - \frac{1}{8 \times det_{\theta^{*}_{i}, \theta_{i}^{'}}\left(\bar{m}\right)} \bigg\} \leq \frac{1}{KT^{2}}
\end{equation}
Thus, we have a finite $\tilde{m}$ that satisfies the constraints of optimization problem~(\ref{opt_prob_dcto}) for $p=1/KT^{2}$, and by definition $m^{*}\left(\frac{1}{KT^{2}}\right) \leq \tilde{m}$. i.e., $m^{*}\left(\frac{1}{KT^{2}}\right)$ is finite.

Given a rotting model, $\theta_{i}^{*}$ of arm $i$, we term that arm `saturated' if it has been pulled at least ${m^{*}\left(\frac{1}{KT^{2}} ; \theta_{i}^{*}\right)}$ times, which is finite since, by definition, ${m^{*}\left(\frac{1}{KT^{2}}; \theta_{i}^{*}\right) \leq m^{*}\left(\frac{1}{KT^{2}}\right)}$. We assume that once an arm is `saturated', it is truely detected every time step, and omit this assertion from now on (we deal with the misdetection case later). i.e., we assume that once arm $i$ hypothesize its rotting model to be $\hat{\theta}_{i}$ and also has been pulled at least $m^{*}\left(\frac{1}{KT^{2}} ; \theta_{i}^{*}\right)$ times, then $\hat{\theta}_{i} = \theta_{i}^{*}$.

We next bound the number of pulls of different arms, given the number of pulls of some other arm. 
Let $s$ be the first time step for which ${\min_{i \in \left[K\right]}\{N_{i}\left(s\right)\} = \max_{\theta \in \Theta^{*}}\{m^{*}\left(\frac{1}{KT^{2}} ; \theta\right)\}}$. We first note that $s$ is finite since by Assumption~\ref{assm_param_rotting} we have ${\mu\left(n ; \theta\right) \in o\left(1\right)}$, combined with the $\argmax$ rule CTO$_{\textrm{SIM}}$ follows and its tie breaking rule, at some finite time step all arms would be pulled the specified amount of times. By our above assumption, from this point on, all the arms' rotting models are correctly detected. Thus, for any arm ${j}$, $N_{j}\left(s\right)$ can be upper bounded by the solution for,
\begin{equation} 
\begin{split}
\min & \text{ } t_{j} \\
\text{s.t } &  \begin{cases} 
t_{j} \in \mathbb{N} \\
t_{j} \geq \max_{\theta \in \Theta^{*}}{\{m^{*}\left(\frac{1}{KT^{2}} ; \theta\right)\}} \\
\mu\left(t_{j}+1 ; \theta^{*}_{j}\right) \leq \min_{\tilde{\theta} \in \Theta} \left[\mu\left(\max_{\theta \in \Theta^{*}}{\bigg\{m^{*}\left(\frac{1}{KT^{2}} ; \theta\right)\bigg\}} ; \tilde{\theta} \right)\right]
\end{cases}
\end{split}
\label{eq_bound_pulls_j}
\end{equation}
where the above optimization bound characterization holds since: \\
(1) For any arm ${j \in \argmin_{i \in \left[K\right]}\{N_{i}\left(s\right)\}}$, this holds trivially by the explicit constraint ${t_{j} \geq \max_{\theta \in \Theta^{*}}{\{m^{*}\left(\frac{1}{KT^{2}} ; \theta\right)\}}}$. \\
(2) For any arm ${j \notin \argmin_{i \in \left[K\right]}\{N_{i}\left(s\right)\}}$, clearly the constraint on the lower bound holds. As for the constraint on the upper bound, it holds by noting that all the arms' hypothesized models are correct and CTO$_{\textrm{SIM}}$ follows an $\argmax$ policy, thus $j$ would not be pulled such that ${\mu\left(N_{j}\left(s\right) ; \theta_{j}^{*}\right) < \min_{\theta \in \Theta}\left[\mu\left(\max_{\theta \in \Theta^{*}}\{m^{*}\left(\frac{1}{KT^{2}} ; \theta\right)\}\right)\right]}$, as the RHS is the lowest obtainable expected reward until time step $s$. In addition, since the tie breaking rule is least \# of pulls, its expected reward would not be equal to ${\min_{\theta \in \Theta}\left[\mu\left(\max_{\theta \in \Theta^{*}}\{m^{*}\left(\frac{1}{KT^{2}} ; \theta\right)\}\right)\right]}$. 

Let ${\mu_{min}\left(s ; \Theta^{*}\right) = \min_{j \in \left[K\right]} \{\mu\left(N_{j}\left(s\right) ; \theta_{j}^{*}\right)\}}$. Following \CTO$_{\textrm{SIM}}$ policy we infer that there exists ${\tilde{s} \geq s}$ for which:\\
(1) ${\mu\left(N_{i}\left(\tilde{s}\right)+1 ; \theta_{i}^{*}\right) \leq \mu_{min}\left(s ; \Theta^{*}\right)}$, for all ${i \in \left[K\right]}$.\\
(2) ${\mu\left(N_{i}\left(\tilde{s}\right) ; \theta_{i}^{*}\right) > \mu_{min}\left(s;\Theta^{*}\right)}$, for all ${i \notin \argmin_{j \in \left[K\right]}\{\mu\left(N_{j}\left(s\right) ; \theta_{j}^{*}\right)\}}$. \\
The above observation holds by noting that \CTO$_{\textrm{SIM}}$ follows an $\argmax$ rule, thus it would choose arms ${\notin \argmin_{j \left[K\right]}\{\mu\left(N_{j}\left(s\right) ; \theta_{j}^{*}\right)\}}$ to be pulled as long as their expected reward is strictly greater than already pulled minimal expected reward ${\mu_{min}\left(s ; \Theta^{*}\right)}$, before the possibility of choosing arms with expected reward ${\leq \mu_{min}\left(s ; \Theta^{*}\right)}$. Since by Eq.~(\ref{eq_bound_pulls_j}) we have that ${\min_{j \in \left[K\right]} \{\mu\left(N_{j}\left(s\right) ; \theta_{j}^{*}\right)\} \geq \min_{\tilde{\theta} \in \Theta} \left[\mu\left(\max_{\theta \in \Theta^{*}}{\{m^{*}\left(\frac{1}{KT^{2}} ; \theta\right)\}} ; \tilde{\theta} \right)\right]}$, we can upper bound $\tilde{s}$ by the following,
\begin{equation} 
\begin{split}
\min & \text{ } \|t\|_{1} \\
\text{s.t } &  \begin{cases} 
t \in \mathbb{N}^{K} \\
t_{i} \geq \max_{\theta \in \Theta^{*}}{\{m^{*}\left(\frac{1}{KT^{2}} ; \theta\right)\}}, \quad \forall i \in \left[K\right] \\
\mu\left(t_{i}+1 ; \theta^{*}_{i}\right) \leq \min_{\tilde{\theta} \in \Theta} \left[\mu\left(\max_{\theta \in \Theta^{*}}{\bigg\{m^{*}\left(\frac{1}{KT^{2}} ; \theta\right)\bigg\}} ; \tilde{\theta} \right)\right], \quad \forall i \in \left[K\right]
\end{cases}
\end{split}
\end{equation}

We turn to show optimality starting from time step $\tilde{s}$. We start by showing for $\tilde{s}$.\\
Assume on the contrary that, ${J\left(\tilde{s} ; \pi^{\textrm{max}}\right) \neq J\left(\tilde{s} ; \pi^{\textrm{\CTO}_{\textrm{SIM}}}\right)}$. On the one hand, by Lemma~\ref{lemma_opt_policy}, we have, $J\left(\tilde{s} ; \pi^{\textrm{max}}\right) \geq J\left(\tilde{s} ; \pi^{\textrm{\CTO}_{\textrm{SIM}}}\right)$. On the other hand, Let $\{q_{i}\}_{i\in\left[K\right]}$ be the set of the arms' number of pulls at time $\tilde{s}$ following $\pi^{\textrm{max}}$ (respectively, $\{\tilde{s}_{i}\}_{i\in\left[K\right]}$ for \CTO$_{\textrm{SIM}}$), i.e., 
\begin{equation}
J\left(\tilde{s} ; \pi^{\textrm{max}}\right) = \sum_{i\in\left[K\right]} \sum_{j=1}^{q_{i}} \mu\left(j ; \theta^{*}_{i}\right)
\end{equation}
We have that $J\left(\tilde{s} ; \pi^{\textrm{\CTO}_{\textrm{SIM}}}\right) - J\left(\tilde{s} ; \pi^{\textrm{max}}\right)$ is a sum of pairs in the form of, $\mu\left(l ; \theta^{*}_{i}\right) - \mu\left(h ; \theta^{*}_{j}\right)$ where $l\leq \tilde{s}_{i}$, and $h>\tilde{s}_{j}$, for $i\neq j\in\left[K\right]$. By definition of $\{\tilde{s}_{i}\}$ and the non-increasing assumption~\ref{assm_rotting}, we have that ${\mu\left(l ; \theta^{*}_{i}\right) \geq \mu_{min}\left(s ; \Theta^{*}\right)}$, and ${\mu_{min}\left(s ; \Theta^{*}\right) \geq \mu\left(h ; \theta^{*}_{j}\right)}$, resulting in ${J\left(\tilde{s} ; \pi^{\textrm{\CTO}_{\textrm{SIM}}}\right) \geq J\left(\tilde{s} ; \pi^{\textrm{max}}\right)}$. Hence, the regret vanishes in time step $\tilde{s}$, achieving optimality.

We next show that the regret remains zero for ${\hat{s} \geq \tilde{s}}$.\\
We showed optimality for time step $\tilde{s}$ defined above. We next show optimality for $\tilde{s}+1$. We examine the two possible cases. \\
\underline{\textit{Case 1}:} $\forall i \in \left[K\right] : q_{i} = \tilde{s}_{i}$. Since \CTO$_{\textrm{SIM}}$ follows the $\argmax$ rule as $\pi^{\textrm{max}}$ does, we infer that arms with equal expected reward would be chosen by both \CTO$_{\textrm{SIM}}$ and $\pi^{\textrm{max}}$. Thereby, holding $J\left(\tilde{s}+1 ; \pi^{\textrm{max}}\right) = J\left(\tilde{s}+1 ; \pi^{\textrm{\CTO}_{\textrm{SIM}}}\right)$. i.e., zero regret as stated. \\
\underline{\textit{Case 2}:} $\exists i : \tilde{s}_{i} \neq q_{i}$. Therefore, there is an arm, denoted as $i_{gap}$, for which ${\tilde{s}_{i_{gap}} < q_{i_{gap}}}$. By the $\argmax$ rule, \CTO$_{\textrm{SIM}}$ chooses an arm $i_{\tilde{s}+1}$ such that, ${\mu\left(\tilde{s}_{i_{\tilde{s}+1}}+1 ; \theta^{*}_{i_{\tilde{s}+1}}\right) \geq \mu\left(\tilde{s}_{i_{gap}}+1 ; \theta^{*}_{i_{gap}}\right)}$. By the non-increasing assumption~\ref{assm_rotting}, and the definition of $\pi^{\textrm{max}}$, since $q_{i_{gap}} \geq \tilde{s}_{i_{gap}} + 1$, we have ${\mu\left(q_{j_{\tilde{s}+1}} ; \theta^{*}_{j_{\tilde{s}+1}}\right) \leq \mu\left(q_{i_{gap}} ; \theta^{*}_{i_{gap}}\right) \leq \mu\left(\tilde{s}_{i_{gap}} + 1 ; \theta^{*}_{i_{gap}}\right)}$, where $j_{\tilde{s}+1}$ is the arm chosen by $\pi^{\textrm{max}}$. Thus, on the one hand we have $J\left(\tilde{s}+1 ; \pi^{\textrm{max}}\right) \leq J\left(\tilde{s}+1 ; \pi^{\textrm{\CTO}_{\textrm{SIM}}}\right)$. On the other hand, by Lemma~\ref{lemma_opt_policy}, we have $J\left(\tilde{s}+1 ; \pi^{\textrm{max}}\right) \geq J\left(\tilde{s}+1 ; \pi^{\textrm{\CTO}_{\textrm{SIM}}}\right)$. Combining the two, we have $J\left(\tilde{s}+1 ; \pi^{\textrm{max}}\right) = J\left(\tilde{s}+1 ; \pi^{\textrm{\CTO}_{\textrm{SIM}}}\right)$. i.e., zero regret as stated. \\
The above argument can be applied recursively for any $\hat{s}>\tilde{s}$, thus establishing optimality of \CTO$_{\textrm{SIM}}$ for all ${\hat{s} \geq s}$, under true detection.

If it happens to be that ${\|t\|_{1} \leq T}$, then for that $T$, \CTO$_{\textrm{SIM}}$ will achieve zero regret (starting from $\tilde{s}$). Since we require that the result will hold from some $T^{*}_{\textrm{SIM}}$ onward, we need the above characterization to also hold for any $\tilde{T} \geq T$. We thereby infer that the smallest $T$ such that for any $\tilde{T} \geq T$, there exists $t$ for which the above stated result holds (i.e., the solution to the optimization problem is indeed holds ${\|t\|_{1} \leq \tilde{T}}$), can serve as an upper bound for $T^{*}_{\textrm{SIM}}$, resulting in $T^{*}_{\textrm{SIM}}$ being upper bounded by the solution for,
\begin{equation}
\begin{split}
\min & \text{ } T \\
\text{s.t } &  \begin{cases} 
T,b \in \mathbb{N}\cup{\{0\}}, t \in \mathbb{N}^{K} \\
\forall b, \exists t : \begin{cases}
\|t\|_{1} \leq T+b \\
t_{i} \geq \max_{\theta \in \Theta^{*}}{\bigg\{m^{*}\left(\frac{1}{K\left(T+b\right)^{2}} ; \theta\right)\bigg\}} \\
\mu\left(t_{i}+1 ; \theta^{*}_{i}\right) \leq  \min_{\tilde{\theta} \in \Theta} \left[\mu\left(\max_{\theta \in \Theta^{*}}{\bigg\{m^{*}\left(\frac{1}{K\left(T+b\right)^{2}} ; \theta\right)\bigg\}} ; \tilde{\theta}\right) \right]
\end{cases}
\end{cases}
\end{split}
\end{equation}
\\
\textbf{Feasibility} \\
In order to show feasibility, we wish to obtain,
\begin{equation} \nonumber
\text{\{\# of steps for Detection\} + \{\# of steps for Balance\}} \leq T
\end{equation}
where Detection is a phase of pulling arms until the rotting models are detected with high enough probability (defined below), and Balance is a phase which at the end of it there is no arm which yields strictly higher expected reward than the minimal observed expected reward so far, as explained in the former step, resulting in vanishing regret (similar to $s$ and $\tilde{s}$ discussed above). We require that the detection of each arm is w.p of at least $1-\frac{1}{KT^{2}}$. Define ${W\left(T\right) = \max_{\theta_{1}, \theta_{2}} \bigg\{det_{\theta_{1}, \theta_{2}}^{\star \downarrow}\left(\frac{1}{16}  \ln^{-1}\left(\sqrt{2K}T\right)\right)\bigg\}}$. As shown in the beginning of this proof, after pulling an arm for ${W\left(T\right)}$ times, the probability of misdetection its rotting model ${\leq \frac{1}{K T^{2}}}$.
We refer to an arm that has been pulled at least ${W\left(T\right)}$ times as `strongly saturated'. From now on we will assume that any `strongly saturated' arm is truely detected at each decision point, and will discuss the other case later on.

On the one hand, by the definition of $bal\left(\right)$, the non-increasing assumption~\ref{assm_rotting}, and the rule of tie breaking applied by \CTO$_{\textrm{SIM}}$, we have that all arms become `strongly saturated' after, at most, ${W\left(T\right) + \left(K-1\right) \times bal\left(W\left(T\right)\right)}$ time steps. \\
On the other hand, from the definition of ${bal\left(\right)}$, and \CTO$_{\textrm{SIM}}$, we infer that no arm would be pulled ${bal\left(W\left(T\right)\right)+1}$ times before all other arms would become `strongly saturated'. \\
Combining the two above observations we have that, after at most ${W\left(T\right) + \left(K-1\right) \times bal\left(W\left(T\right)\right)}$ time steps, there exists a time step in which all arms have became `strongly saturated', but were not pulled more than ${bal\left(W\left(T\right)\right)}$ times. From that point, following the same flow at the former subsection, the total number of pulls required in order to ``balance" the arms (i.e., there is no pull that would yield strictly higher reward than the minimal expected reward observed so far), is bounded by ${K \times bal\left(W\left(T\right)\right)}$. That is under the worst case scenario, where every arm that becomes `strongly saturated` is detected to be an arm that requires ${bal\left(W\left(T\right)\right)}$ pulls to ``balance" itself w.r.t to another `strongly saturated' arm. Thus, we infer that,
\begin{equation} \nonumber
\text{\{\# of steps for Detection\} + \{\# of steps for Balance\}} \leq K\times bal\left(W\left(T\right)\right)
\end{equation}
Let $\epsilon = \left(K\sqrt{2K}\right)^{-1}$.
By assumption~\ref{assm_sim}, we have that there exists a finite $\tilde{T}_{max}$ for which,
\begin{equation} 
\forall \tilde{T} \geq \tilde{T}_{max} : bal\left(\max_{\theta_{1} \neq \theta_{2} \in \Theta^{2}}\bigg\{det_{\theta_{1}, \theta_{2}}^{\star \downarrow}\left(\frac{1}{16}  \ln^{-1}\left(\tilde{T}\right)\right)\bigg\} \right) \leq \epsilon \tilde{T}
\end{equation}
We denote $T = \left(\sqrt{2K}\right)^{-1}\tilde{T}$, and get,
\begin{equation} 
\forall T \geq \frac{\tilde{T}_{max}}{\sqrt{2K}} : K\times bal\left(W\left(T\right)\right) \leq T
\end{equation}
which implies, under true detection, that $\forall T \geq \tilde{T}_{max} / \sqrt{2K}$, \CTO$_{\textrm{SIM}}$ algorithm achieves zero regret. \\
Since by definition we have ${\forall \theta \in \Theta : m^{*}\left(\frac{1}{KT^{2}} ; \theta\right) \leq m^{*}\left(\frac{1}{KT^{2}}\right)}$, and by definition of ${m^{*}\left(\frac{1}{KT^{2}}\right)}$ we have ${m^{*}\left(\frac{1}{KT^{2}}\right) \leq W\left(T\right)}$, we infer that there exists (a finite) $T^{*}_{\textrm{SIM}}$ that holds the optimization problem characterization as stated above (i.e., $\forall \tilde{T} \geq T^{*}_{\textrm{SIM}}$ the optimization problem is feasible).\\
\\
\textbf{Misdetection and Expectation} \\
So far, we assumed that each `saturated' (or `strongly saturated') arm is truely detected. By definition each `saturated' (or `strongly saturated') arm probability of misdetection in any time step is upper bounded by $1/KT^{2}$. Thereby, after all the arms are `saturated', the probability of a misdetection in each time step is upper bounded by $1/T^{2}$. The number of time steps where all the arms are `saturated' (referred to as the `saturated step') is trivially bounded by $T$. Hence, the probability that a misdetection occurs after the `saturated step' is bounded by $1/T$. Meaning that $\forall T \geq T^{*}_{\textrm{SIM}}$, \CTO$_{\textrm{SIM}}$ achieves zero regret w.p of at least $1-1/T$.\\
Next, we note that, as for the case where we misdetect any arm,
\begin{equation} 
\begin{split}
J\left(T ; \pi^{\textrm{max}}\right) - J\left(T ; \pi^{\textrm{CTO}_{\textrm{SIM}}}\right) & = \sum_{i=1}^{K} \sum_{j=1}^{N_{i}^{\textrm{max}}\left(T\right)} \mu\left( j ; \theta^{*}_{i}  \right) - \sum_{i=1}^{K} \sum_{j=1}^{N_{i}^{\textrm{CTO}_{\textrm{SIM}}}\left(T\right)} \mu\left( j ; \theta^{*}_{i}  \right) \\
& \leq \sum_{i=1}^{K} I_{\{N_{i}^{\textrm{max}}\left(T\right) > N_{i}^{\textrm{CTO}_{\textrm{SIM}}}\left(T\right)\}} \sum_{N_{i}^{\textrm{CTO}_{\textrm{SIM}}}\left(T\right)+1}^{N_{i}^{\textrm{max}}\left(T\right)} \mu\left( j ; \theta^{*}_{i}  \right) \\
& \leq T  \max_{\theta \in \Theta^{*}}{\bigg\{\mu\left( \min_{i \in \left[K\right]} \{ N_{i}^{\textrm{\CTO}_{\textrm{SIM}}}\left(T\right) \} ; \theta \right)}\bigg\}
\end{split}
\end{equation}
where the first inequality holds by only considering cases where $N_{i}^{\textrm{max}}\left(T\right) > N_{i}^{\textrm{CTO}_{\textrm{SIM}}}\left(T\right)$, and not the other way around (since the expected rewards are positive by Assumption~\ref{assm_param_rotting}).

By applying expectation over events (true detection or not), we get, 
\begin{equation} 
\begin{split}
\mathcal{R}\left(T ; \pi^{\textrm{CTO}_{\textrm{SIM}}} \right) & = \mathcal{R}\left(T ; \pi^{\textrm{CTO}_{\textrm{SIM}}} | \text{true detection} \right) \times P\left(\text{true detection}\right) \\
& \qquad + \mathcal{R}\left(T ; \pi^{\textrm{CTO}_{\textrm{SIM}}} | \text{misdetection} \right) \times P\left(\text{misdetection}\right) \\
& \leq \max_{\theta \in \Theta^{*}}{\bigg\{\mu\left( \min_{i \in \left[K\right]} \{ N_{i}^{\textrm{\CTO}_{\textrm{SIM}}}\left(T\right) \} ; \theta \right)}\bigg\}
\end{split}
\end{equation}

Finally,
\begin{equation}
\begin{split}
T & = \sum_{i=1}^{K} N_{i}^{\textrm{\CTO}_{\textrm{SIM}}}\left(T\right) \\
& \leq \min_{i\in\left[K\right]} N_{i}^{\textrm{\CTO}_{\textrm{SIM}}}\left(T\right) + \left(K-1\right)  \max_{i\in\left[K\right]} N_{i}^{\textrm{\CTO}_{\textrm{SIM}}}\left(T\right) \\
& \leq \min_{i\in\left[K\right]} N_{i}^{\textrm{\CTO}_{\textrm{SIM}}}\left(T\right) + \left(K-1\right) \times bal\left(\min_{i\in\left[K\right]} N_{i}^{\textrm{\CTO}_{\textrm{SIM}}}\left(T\right)\right) \\
& \leq K \times bal\left(\min_{i\in\left[K\right]} N_{i}^{\textrm{\CTO}_{\textrm{SIM}}}\left(T\right)\right)
\end{split}
\end{equation}
Hence, by assumption~\ref{assm_rotting}, ${\min_{i \in \left[K\right]}N_{i}^{\textrm{\CTO}_{\textrm{SIM}}}\left(T\right)\overset{T \rightarrow \infty}{\longrightarrow} \infty}$, resulting in ${\mathcal{R}\left(T ; \pi^{\textrm{CTO}_{\textrm{SIM}}} \right) \in o\left(1\right)}$, and trivially $\leq \max_{\theta \in \Theta^{*}} \mu\left(1;\theta\right)$.

We \textbf{Note} that from the feasibility step, given a function $U\left(\epsilon\right)$ that satisfies ${\forall n \geq U\left(\epsilon\right)}$,
\begin{equation} 
bal\left(\max_{\theta_{1} \neq \theta_{2} \in \Theta^{2}} \bigg\{ det^{\star \downarrow}_{\theta_{1},\theta_{2}}\left(\frac{1}{16}\ln^{-1}\left(n\right)\right) \bigg\} \right) \leq \epsilon n
\end{equation}
we have,
\begin{equation} 
T^{*}_{\textrm{SIM}} \leq \frac{U\left(\left(K \sqrt{2K}\right)^{-1}\right)}{\sqrt{2K}}
\end{equation}

\subsection{Proof of Thm.~\ref{thm_Bcto}}
\textbf{Decomposing the regret} \\
First, we upper bound the regret by,
\begin{equation}
\begin{split}
\mathcal{R}\left(T ; \pi^{\textrm{D-\CTO}_{\textrm{UCB}}}\right) & = \sum_{i=1}^{K}\sum_{j=1}^{\mathbb{E}\left[N_{i}^{\pi^{\textrm{max}}}\left(T\right)\right]} \mu_{i}\left(j\right) - \sum_{i=1}^{K}\sum_{j=1}^{\mathbb{E}\left[N_{i}^{\pi^{\textrm{D-\CTO}_{\textrm{UCB}}}}\left(T\right)\right]} \mu_{i}\left(j\right) \\
& \leq \underbrace{\sum_{i \neq a^{*}}\sum_{j=1}^{\mu^{\star \downarrow}\left(\Delta_{i} ; \theta_{i}^{*}\right)} \mu_{i}\left(j\right)}_{= \tilde{C}\left(\Theta^{*} , \{\mu_{i}^{c}\}\right)} + \sum_{j=1}^{T} \mu_{a^{*}}\left(j\right) - \sum_{i=1}^{K}\sum_{j=1}^{\mathbb{E}\left[N_{i}^{\pi^{\textrm{D-\CTO}_{\textrm{UCB}}}}\left(T\right)\right]} \mu_{i}\left(j\right) \\
& = \tilde{C}\left(\Theta^{*} , \{\mu_{i}^{c}\}\right) + \sum_{\mathbb{E}\left[N_{a^{*}}^{\pi^{\textrm{max}}}\left(T\right)\right]+1}^{T} \mu_{a^{*}}\left(j\right) - \sum_{i \neq a^{*}}\sum_{j=1}^{\mathbb{E}\left[N_{i}^{\pi^{\textrm{D-\CTO}_{\textrm{UCB}}}}\left(T\right)\right]} \mu_{i}\left(j\right) \\
& \leq \tilde{C}\left(\Theta^{*} , \{\mu_{i}^{c}\}\right) + \sum_{\mathbb{E}\left[N_{a^{*}}^{\pi^{\textrm{max}}}\left(T\right)\right]+1}^{T}\left(\mu_{a^{*}}^{c} + \mu\left(1 ; \theta_{a^{*}}^{*}\right)\right) - \sum_{i \neq a^{*}}\sum_{j=1}^{\mathbb{E}\left[N_{i}^{\pi^{\textrm{D-\CTO}_{\textrm{UCB}}}}\left(T\right)\right]} \mu_{i}^{c} \\
& \leq \tilde{C}\left(\Theta^{*} , \{\mu_{i}^{c}\}\right) + \sum_{i \neq a^{*}} \mathbb{E}\left[N_{i}^{\pi^{\textrm{D-\CTO}_{\textrm{UCB}}}}\left(T\right)\right] \times \left(\Delta_{i} + \mu\left(1 ; \theta_{a^{*}}^{*}\right)\right)
\end{split}
\end{equation}
where $\mathbb{E}\left[N_{i}^{\pi^{\textrm{max}}}\left(T\right)\right]$ is the expected number of pulls of arm $i$ at time $T$ induced by the optimal policy, $\pi^{\textrm{max}}$, and $\mathbb{E}N_{i}^{\pi^{\textrm{D-\CTO}_{\textrm{UCB}}}}\left(T\right)$ is the expected number of pulls induced by policy $\pi^{\textrm{D-\CTO}_{\textrm{UCB}}}$. The first inequality holds by noting that $\pi^{\textrm{max}}$ pulls according to $\argmax$ rule, thus any arm $i \neq a^{*}$ would not be pulled after yielding expected reward not greater than $\mu_{a^{*}}^{c}$, according to the behavior of $\mu\left(\cdot;\cdot\right)$ by assumption~\ref{assm_rotting}. \\
\\
\textbf{Detecting the models} \\
Next, we show that $m^{*}_{\text{diff}}\left(\delta/K\right)$ is finite. Define,
\begin{equation}
D\left(\mu\left(\cdot;\theta\right),1,n\right) = \sum_{j=1}^{\floor{\frac{n}{2}}} \mu\left(j ; \theta\right) - \sum_{j=\floor{\frac{n}{2}}+1}^{n} \mu\left(j ; \theta\right)
\end{equation}
and,
\begin{equation} 
\theta_{i}^{'}\left(\tilde{m}\right) = \argmin_{\theta \neq \theta_{i}^{*}} \bigg\{ \bigg| \mathcal{D}\left(\mu\left(\cdot ; \theta_{i}^{*}\right), 1, \tilde{m}\right) -  \mathcal{D}\left(\mu\left(\cdot ; \theta\right), 1, \tilde{m}\right) \bigg| \bigg\}
\end{equation}
Thus, we have, when we sample only from arm $i$, and for an even $\tilde{m}$
\begin{equation} 
\begin{split}
P\left( \hat{\theta}_{i}\left(\tilde{m}\right) \neq \theta^{*}_{i} \right) & =  P\left( \exists \theta \neq \theta^{*}_{i} : |Z\left(i,\tilde{m}; \theta\right)| \leq |Z\left(i ,\tilde{m}; \theta^{*}_{i}\right)| \right) \\
& \leq P\left( \bigg| \left( \sum_{j=1}^{\frac{\tilde{m}}{2}}r_{j}^{i} - \sum_{j=\frac{\tilde{m}}{2}+1}^{\tilde{m}}r_{j}^{i} \right) - \mathcal{D}\left(\mu\left(\cdot ; \theta_{i}^{*}\right), 1, \tilde{m}\right) \bigg| > \right. \\
& \qquad \left. \frac{1}{2}  \bigg| \mathcal{D}\left(\mu\left(\cdot ; \theta_{i}^{*}\right), 1, \tilde{m}\right) -  \mathcal{D}\left(\mu\left(\cdot ; \theta_{i}^{'}\left(\tilde{m}\right)\right), 1, \tilde{m}\right) \bigg| \right)  \\
& \leq 2  \exp \bigg\{ - \frac{1}{8 \times Ddet_{\theta^{*}_{i}, \theta_{i}^{'}\left(\tilde{m}\right)}\left(\tilde{m}\right)} \bigg\}
\end{split}
\end{equation}
where the first inequality holds by inclusion of events, and the second inequality holds by Eq.~(\ref{hoeff_ineq}), the definition of $Ddet_{\theta_{i}^{*},\theta_{i}^{'}}$, and noting that for an even $\tilde{m}$ we have,
\begin{equation} 
\mathbb{E} \left[ \sum_{j=1}^{\frac{\tilde{m}}{2}}r_{j}^{i} - \sum_{j=\frac{\tilde{m}}{2}+1}^{\tilde{m}}r_{j}^{i} \right] = \mathcal{D}\left(\mu\left(\cdot ; \theta_{i}^{*}\right), 1, \tilde{m}\right)
\end{equation}
By assumption~\ref{assm_Bdistinction}, there exists a finite, even, $\tilde{m}$ for which,
\begin{equation} 
\max_{\theta_{1} \neq \theta_{2} \in \Theta^{2}} \bigg\{ Ddet_{\theta_{1}, \theta_{2}} \left(\tilde{m}\right) \bigg\} \leq \frac{1}{8}  \ln ^{-1} \left(\frac{2K}{\delta}\right)
\end{equation}
If we plug $\tilde{m}$ back to the above equation we get,
\begin{equation}
2  \exp \bigg\{ - \frac{1}{8 \times Ddet_{\theta^{*}_{i}, \theta_{i}^{'}\left(\tilde{m}\right)}\left(\tilde{m}\right)} \bigg\} \leq \frac{\delta}{K}
\end{equation}
Thus, we have a finite $\tilde{m}$ that satisfies the constraints of Prob.~(\ref{opt_prob_dcto}) for $p=\delta/K$, and by definition ${m^{*}_{\text{diff}}\left(\delta/K\right) \leq \tilde{m}}$. i.e., $m^{*}_{\text{diff}}\left(\delta/K\right)$ is finite.\\
\\
\textbf{Bounding number of pulls} \\
We wish to bound $\mathbb{E}\left[N_{i}^{\pi^{\textrm{D-\CTO}_{\textrm{UCB}}}}\left(T\right)\right]$ for all $i \neq a^{*}$. Remember that in the exploration part (leading to the Detect step), we pull each arm $m^{*}_{\text{diff}}\left(\delta/K\right)$ times, hence,
\begin{equation}
N_{i}^{\pi^{\textrm{D-\CTO}_{\textrm{UCB}}}}\left(T\right) = m^{*}_{\text{diff}}\left(\delta/K\right) + \sum_{t=K\times m^{*}_{\text{diff}}\left(\delta/K\right)+1}^{T} 1_{\{i\left(t\right)=i\}}
\end{equation}
where $1_{\{\cdot\}}$ is the indicator function. Similarly to the proof of UCB1 (\cite{FiniteTimeAnalysisMAB}) we have,
\begin{equation} 
N_{i}^{\pi^{\textrm{D-\CTO}_{\textrm{UCB}}}}\left(T\right) \leq l_{i} + \sum_{t=1}^{\infty} \sum_{s=m^{*}_{\text{diff}}\left(\delta/K\right)}^{t-1} \sum_{s_{i}=l_{i}}^{t-1} 1_{\{\hat{\mu}_{a^{*}}^{c}\left(s\right)+\mu\left(s;\theta^{*}_{a^{*}}\right)+c_{t,s} \leq \hat{\mu}_{i}^{c}\left(s_{i}\right)+\mu\left(s_{i};\theta^{*}_{i}\right)+c_{t,s_{i}}\}}
\end{equation}
where for some $\epsilon_{i} \in \left(0, \Delta_{i}\right)$, we denote ${l_{i} = \max \bigg\{ m^{*}_{\text{diff}}\left(\delta/K\right), \mu^{\star \downarrow}\left(\epsilon_{i} ; \theta_{i}^{*}\right), \ceil{\frac{32 \sigma^{2} \ln T}{\left(\Delta_{i} - \epsilon_{i}\right)^{2}}} \bigg\}}$, and we note that we assume that we have detected the true underlying rotting models (holds w.p of at least $1-\delta$ as shown above). \\
The above indicator function holds when at least one of the following holds,
\begin{equation} 
\begin{cases}
\hat{\mu}^{c}_{a^{*}}\left(s\right) \leq \mu_{a^{*}}^{c} - c_{t,s} \\
\hat{\mu}^{c}_{i}\left(s_{i}\right) \geq \mu_{i}^{c} + c_{t,s_{i}} \\
\mu_{a^{*}}^{c} + \mu\left(s ; \theta_{a^{*}}^{*}\right) < \mu_{i}^{c} + \mu\left(s_{i} ; \theta_{i}^{*}\right) + 2 c_{t,s_{i}}
\end{cases}
\end{equation}
Plugging $c_{t,s}$ and $c_{t,s_{i}}$, and using Eq.~(\ref{hoeff_ineq}), we have,
\begin{equation} 
\begin{cases}
P\left(\hat{\mu}^{c}_{a^{*}}\left(s\right) \leq \mu_{a^{*}}^{c} - c_{t,s}\right) = t^{-4} \\
P\left(\hat{\mu}^{c}_{i}\left(s_{i}\right) \geq \mu_{i}^{c} + c_{t,s_{i}}\right) = t^{-4}
\end{cases}
\end{equation}
And for $s_{i} \geq l_{i}$ we have,
\begin{equation} 
\begin{split}
\mu_{a^{*}}^{c} + \mu\left(s ; \theta_{a^{*}}^{*}\right) - \mu_{i}^{c} - \mu\left(s_{i} ; \theta_{i}^{*}\right) - 2 c_{t,s_{i}} & \geq \mu_{a^{*}}^{c} - \mu_{i}^{c} - \mu\left(s_{i} ; \theta_{i}^{*}\right) - 2 c_{t,s_{i}} \\
& \geq \mu_{a^{*}}^{c} - \mu_{i}^{c} - \epsilon_{i} - 2 c_{t,s_{i}} \\
& = \left(\Delta_{i} - \epsilon_{i}\right) - 2c_{t,s_{i}} \\
& \geq 0
\end{split}
\end{equation}
where the first inequality holds by assumption~\ref{assm_param_rotting}, the second inequality by ${s_{i} \geq \mu^{\star \downarrow}\left(\epsilon_{i} ; \theta_{i}^{*}\right)}$, and the third inequality by ${s_{i} \geq  \ceil{\frac{32 \sigma^{2} \ln T}{\left(\Delta_{i} - \epsilon_{i}\right)^{2}}}}$.\\
Thus, combining the above observations, we get,\
\begin{equation} 
\begin{split}
\mathbb{E}\left[N_{i}^{\pi}\left(T\right)\right] & \leq l_{i} + \sum_{t=1}^{\infty} \sum_{s=m^{*}_{\text{diff}}\left(\delta/K\right)}^{t-1} \sum_{s_{i}=l_{i}}^{t-1} \left( P\left(\hat{\mu}^{c}_{a^{*}} \leq \mu_{a^{*}}^{c} - c_{t,s}\right) +  P\left(\hat{\mu}^{c}_{i} \geq \mu_{i}^{c} + c_{t,s_{i}}\right)\right) \\
& \leq l_{i} + \frac{\pi^{2}}{3}
\end{split}
\end{equation}
Denoting ${C\left(\Theta^{*}, \{\mu_{i}^{c}\}\right) = \tilde{C}\left(\Theta^{*}, \{\mu_{i}^{c}\}\right) + \sum_{i \neq a^{*}} \frac{\pi^{2}+3}{3}  \left(\Delta_{i} + \mu\left(1 ; \theta_{a^{*}}^{*}\right)\right)}$, and plugging back into the upper bound on the regret, we achieve the stated result.

\section{Example~\ref{expl}}
Next, we show an example for which the different assumptions hold; the case where the reward of arm $i$ for its $n^{\textrm{th}}$ pull is distributed as ${\mathcal{N}\left(\mu_{i}^{c}+n^{-\theta_{i}^{*}}, \sigma^{2}\right)}$. Where ${\theta_{i}^{*} \in \Theta = \{\theta_{1},\theta_{2},...,\theta_{M}\}}$, and ${\forall \theta \in \Theta : 0.01 \leq\theta \leq 0.49}$.

\subsection{Assumption~\ref{assm_param_rotting}}
The assumption given by $\mu\left(n ; \theta\right)$ is positive, non-increasing in $n$, and 	${\mu\left(n ; \theta\right) \in o\left(1\right), \forall \theta \in \Theta}$, where $\Theta$ is a discrete known set. Indeed, for any ${\theta \in \{\theta_{1}, \theta_{2},...,\theta_{M}\}}$, which is a discrete known set where ${0.01 \leq\theta \leq 0.49}$, we have ${n^{-\theta} \geq 0}$ for all ${n \geq 1}$. Moreover, ${\frac{\partial n^{-\theta}}{\partial \theta} = -\theta n^{-\theta-1} < 0}$ for all ${n \geq 1}$, and ${n^{-\theta} \overset{n \rightarrow \infty}{\longrightarrow} 0}$.

\subsection{Assumption~\ref{assm_sim}}
The assumption is given by,
\begin{equation}
bal\left(\max_{\theta_{1} \neq \theta_{2} \in \Theta^{2}} \bigg\{det_{\theta_{1}, \theta_{2}}^{\star \downarrow}\left(\frac{1}{16}  \ln^{-1}\left(\zeta\right)\right)\bigg\}\right) \in o \left(\zeta\right)
\end{equation}
Without a loss of generality, assume ${\theta_{2} > \theta_{1}}$. We have for large enough $n$,
\begin{equation} 
\begin{split}
det_{\theta_{1}, \theta_{2}}\left(n\right) & = \frac{n\sigma^{2}}{\left(\sum_{j=1}^{n} j^{-\theta_{1}} - \sum_{j=1}^{n} j^{-\theta_{2}}\right)^{2}} \\
& \leq \frac{n\sigma^{2}}{\left(c_{1}n^{1-\theta_{1}}-c_{1}-c_{2}n^{1-\theta_{2}}\right)^{2}} \\
& = \frac{n\sigma^{2}}{c_{1}^{2}n^{2-2\theta_{1}}+c_{2}^{2}n^{2-2\theta_{2}}-2c_{1}c_{2}n^{2-\theta_{1}-\theta_{2}}-2c_{1}^{2}n^{1-\theta_{1}}+2c_{1}c_{2}n^{1-\theta_{2}}+c_{1}^{2}} \\
& \leq \frac{n\sigma^{2}}{\tilde{c}n^{2-2\theta_{1}}} \\
& = \frac{\bar{c}}{n^{1-2\theta_{1}}}
\end{split}
\end{equation}
where ${\{c_{1}, c_{2}, \tilde{c}, \bar{c}\}}$ are positive constants (independent of $n$). The first inequality holds by bounding the sums by integrals and keeping in mind that ${\theta_{2} > \theta_{1}}$ combined with $0.01 \leq \theta \leq 0.49$. The second inequality holds from large enough $n$ (leading exponent, depends only on ${\{\theta_{1}, \theta_{2}\}}$, but finite). \\
Next, we have,
\begin{equation}
\frac{\bar{c}}{n^{1-2\theta_{1}}} < \frac{1}{16}\ln^{-1}\left(\zeta\right) \Longrightarrow n > \left(16\bar{c}\ln\left(\zeta\right)\right)^{\frac{1}{1-2\theta_{1}}} > \left(16\bar{c}\ln\left(\zeta\right)\right)^{50}
\end{equation}
Meaning that $\zeta$ large enough,
\begin{equation} 
\max_{\theta_{1} \neq \theta_{2} \in \Theta^{2}} \bigg\{det_{\theta_{1}, \theta_{2}}^{\star \downarrow}\left(\frac{1}{16}  \ln^{-1}\left(\zeta\right)\right)\bigg\} < \left(16\bar{c}\ln\left(\zeta\right)\right)^{50}
\end{equation} 
Next, we have,
\begin{equation} 
\alpha^{-0.1} \leq x^{-0.49} \Longrightarrow \alpha \geq x^{4.9}
\end{equation}
Hence, $bal\left(x\right) = x^{4.9}$. Since $bal\left(\cdot\right)$ is monotonically increasing, we have that for $\zeta$ large enough,
\begin{equation}
bal\left(\max_{\theta_{1} \neq \theta_{2} \in \Theta^{2}} \bigg\{det_{\theta_{1}, \theta_{2}}^{\star \downarrow}\left(\frac{1}{16}  \ln^{-1}\left(\zeta\right)\right)\bigg\}\right) < \hat{c}\ln^{245}\left(\zeta\right)
\end{equation}
where $\hat{c}$ is a positive constant (independent of $\zeta$). Finally, we note that,
\begin{equation}
\lim_{\zeta \rightarrow \infty} \frac{\ln^{245}\left(\zeta\right)}{\zeta} = 0
\end{equation}
Thus we infer that the assumption holds.

\subsection{Assumption~\ref{assm_Bdistinction}}
The assumption is given by,
\begin{equation}
\max_{\theta_{1} \neq \theta_{2} \in \Theta^{2}} \bigg\{Ddet^{\star \downarrow}_{\theta_{1}, \theta_{2}}\left(\epsilon\right)\bigg\} \leq B\left(\epsilon\right) < \infty, \quad \forall \epsilon > 0
\end{equation}
Without a loss of generality, assume ${\theta_{2} > \theta_{1}}$. We have for large enough $n$,
\begin{equation} 
\begin{split}
Ddet_{\theta_{1}, \theta_{2}}\left(n\right) & = \frac{n\sigma^{2}}{\left(\left(\sum_{j=1}^{\floor*{\frac{n}{2}}}j^{-\theta_{1}} - \sum_{j=\floor*{\frac{n}{2}}+1}^{n}j^{-\theta_{1}}\right) - \left(\sum_{j=1}^{\floor*{\frac{n}{2}}}j^{-\theta_{2}} - \sum_{j=\floor*{\frac{n}{2}}+1}^{n}j^{-\theta_{2}}\right)\right)^{2}} \\
& \leq \frac{n\sigma^{2}}{\left( c_{1}\left( -1+2\floor*{\frac{n}{2}}^{1-\theta_{1}}-n^{1-\theta_{1}} \right) - c_{2}\left( 2\left(\floor*{\frac{n}{2}}+1\right)^{1-\theta_{2}} -n^{1-\theta_{2}} \right) \right)^{2}} \\
& \leq \frac{n\sigma^{2}}{\tilde{c}n^{2-2\theta_{1}}} \\
& = \frac{\tilde{c}}{n^{1-2\theta_{1}}}
\end{split}
\end{equation}
where ${\{c_{1}, c_{2}, \tilde{c}\}}$ are positive constants (independent of $n$). The inequalities hold by the same arguments as in E.2. Again, following the same logic as the end of E.2, we have that the assumption holds.

\end{document}